%% file: iNat_main_paper.tex
\documentclass[10pt,twocolumn,letterpaper]{article}

\usepackage{cvpr}
\usepackage{times}
\usepackage{epsfig}
\usepackage{graphicx}
\usepackage{amsmath}
\usepackage{amssymb}
\usepackage{cases}
\usepackage[table]{xcolor}

\usepackage[pagebackref=true,breaklinks=true,letterpaper=true,colorlinks,bookmarks=false]{hyperref}

\cvprfinalcopy 
\def\cvprPaperID{1916} 

\ifcvprfinal\fi
\begin{document}

\title{The iNaturalist Species Classification and Detection Dataset}

\author{Grant Van Horn$^{1}$\hspace{15pt}Oisin Mac Aodha$^{1}$\hspace{15pt}Yang Song$^{2}$\hspace{15pt}Yin Cui$^{3}$\hspace{15pt}Chen Sun$^{2}$\\Alex Shepard$^{4}$\hspace{15pt}Hartwig Adam$^{2}$\hspace{15pt}Pietro Perona$^{1}$\hspace{15pt}Serge Belongie$^{3}$\\
\\
$^{1}$Caltech\hspace{20pt}$^{2}$Google\hspace{20pt}$^{3}$Cornell Tech\hspace{20pt}$^{4}$iNaturalist}

\maketitle
\thispagestyle{empty}

\begin{abstract}
Existing image classification datasets used in computer vision tend to have a uniform distribution of images across object categories. 
In contrast, the natural world is heavily imbalanced, as some species are more abundant and easier to photograph than others. 
To encourage further progress in challenging real world conditions we present the iNaturalist species classification and detection dataset, consisting of 859,000 images from over 5,000 different species of plants and animals.
It features visually similar species, captured in a wide variety of situations, from all over the world.
Images were collected with different camera types, have varying image quality, feature a large class imbalance, and have been verified by multiple citizen scientists. We discuss the collection of the dataset and present extensive baseline experiments using state-of-the-art computer vision classification and detection models.
Results show that current non-ensemble based methods achieve only 67\% top one classification accuracy, illustrating the difficulty of the dataset.
Specifically, we observe poor results for classes with small numbers of training examples suggesting more attention is needed in low-shot learning. 

\end{abstract}

%
%
%
\section{Introduction}
Performance on existing image classification benchmarks such as \cite{russakovsky2015imagenet} is close to being saturated by the current generation of classification algorithms \cite{he2016deep, szegedy2016rethinking,szegedy2016inception,xie2016aggregated}. 
However, the number of training images is crucial. 
If one reduces the number of training images per category, typically performance suffers.
It may be tempting to try and acquire more training data for the classes with few images but this is often impractical, or even impossible, in many application domains. 
We argue that class imbalance is a property of the real world and computer vision models should be able to deal with it. 
Motivated by this problem, we introduce the iNaturalist Classification and Detection Dataset (iNat2017).
Just like the real world, it exhibits a large class imbalance, as some species are much more likely to be observed.

\begin{figure}[t]
\centering
\includegraphics[width=0.48\textwidth]{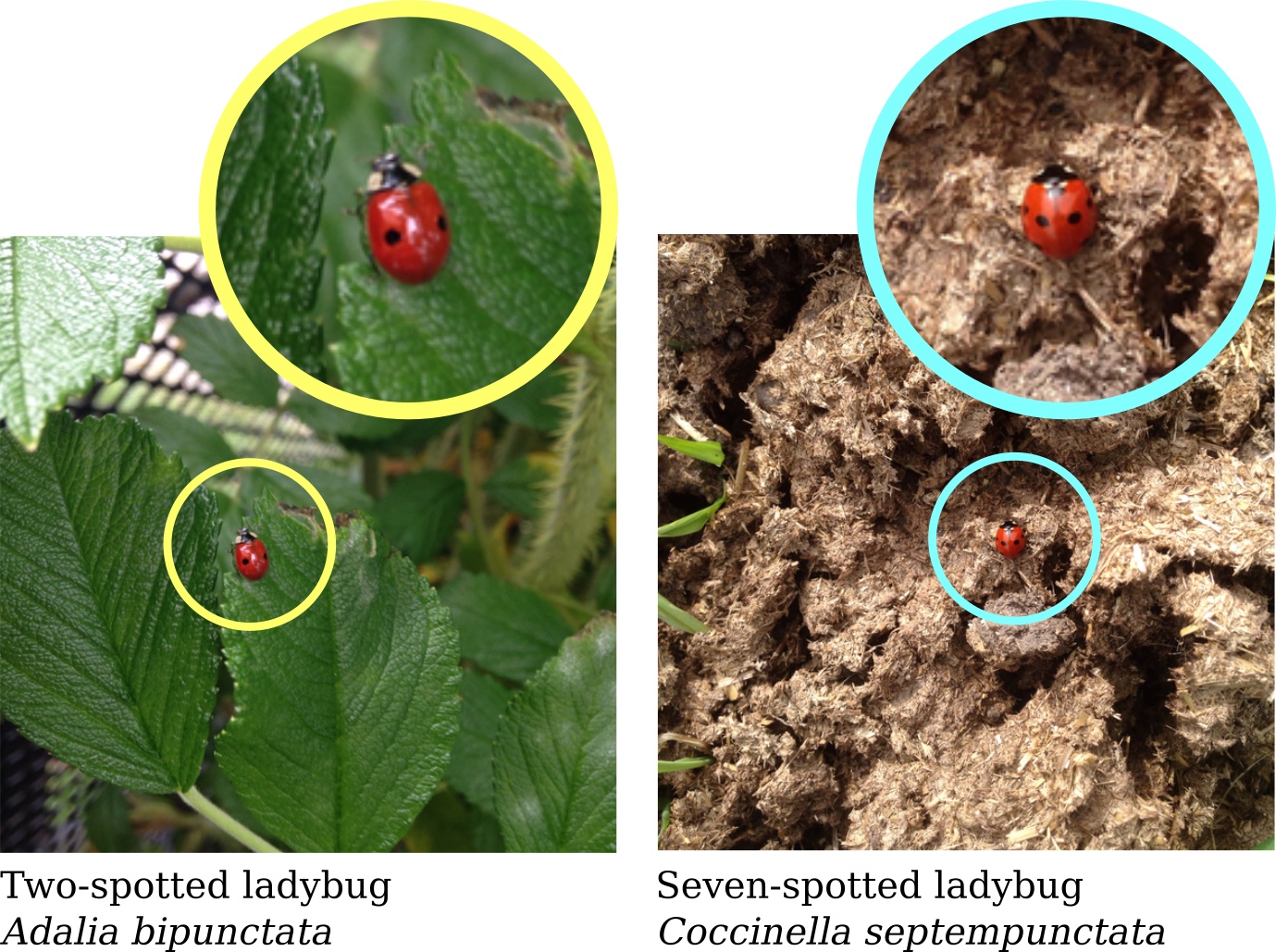}
\caption{Two visually similar species from the iNat2017 dataset. 
Through close inspection, we can see that the ladybug on the left has \emph{two} spots while the one on the right has \emph{seven}.} 
\label{fig:exemplar_res}
\vspace{-10pt}
\end{figure}

It is estimated that the natural world contains several million species with around 1.2 million of these having already been formally described \cite{mora2011many}. 
For some species, it may only be possible to determine the species via genetics or by dissection.
For the rest, visual identification in the wild, while possible, can be extremely challenging. 
This can be due to the sheer number of visually similar categories that an individual would be required to remember along with the challenging inter-class similarity; see Fig.~\ref{fig:exemplar_res}. 
As a result, there is a critical need for robust and accurate automated tools to scale up biodiversity monitoring on a global scale \cite{cardinale2012biodiversity}. 

The iNat2017 dataset is comprised of images and labels from the citizen science website iNaturalist\footnote{\url{www.inaturalist.org}}.
The site allows naturalists to map and share photographic observations of biodiversity across the globe. 
Each observation consists of a date, location, images, and labels containing the name of the species present in the image.
As of November 2017, iNaturalist has collected over 6.6 million observations from 127,000 species. 
From this, there are close to 12,000  species that have been observed by at least twenty people and have had their species ID confirmed by multiple annotators.

The goal of iNat2017 is to push the state-of-the-art in image classification and detection for `in the wild' data featuring large numbers of imbalanced, fine-grained, categories.
iNat2017 contains over 5,000 species, with a combined training and validation set of 675,000 images, 183,000 test images, and over 560,000 manually created bounding boxes.
It is free from one of the main selection biases that are encountered in many existing computer vision datasets - as opposed to being scraped from the web all images have been collected and then verified by multiple citizen scientists. 
It features many visually similar species, captured in a wide variety of situations, from all over the world. 
We outline how the dataset was collected and report extensive baseline performance for state-of-the-art classification and detection algorithms.
Our results indicate that iNat2017 is challenging for current models due to its imbalanced nature and will serve as a good experimental platform for future advances in our field.

\section{Related Datasets}
In this section we review existing image classification datasets commonly used in computer vision. 
Our focus is on large scale, fine-grained, object categories as opposed to datasets that feature common everyday objects, \eg~\cite{fei2007learning, everingham2010pascal, lin2014microsoft}.
Fine-grained classification problems typically exhibit two distinguishing differences from their coarse grained counter parts. First, there tends to be only a small number of domain experts that are capable of making the classifications. Second, as we move down the spectrum of granularity, the number of instances in each class becomes smaller. 
This motivates the need for automated systems that are capable of discriminating between large numbers of potentially visually similar categories with small numbers of training examples for some categories. 
In the extreme, face identification can be viewed as an instance of fine-grained classification and many existing benchmark datasets with long tail distributions exist \eg \cite{LFWTech, parkhi2015deep, guo2016ms, cao2017vggface2}. 
However, due to the underlying geometric similarity between faces, current state-of-the-art approaches for face identification tend to perform a large amount of face specific pre-processing \cite{taigman2014deepface, schroff2015facenet, parkhi2015deep}.

The vision community has released many fine-grained datasets covering several domains such as birds \cite{welinder2010caltech,wah2011caltech,berg2014birdsnap,van2015building, krause2016unreasonable}, dogs \cite{KhoslaYaoJayadevaprakashFeiFei_FGVC2011,parkhi12a,liu2012dog}, airplanes \cite{maji2013fine,vedaldi2014understanding}, flowers \cite{nilsback2006visual}, leaves \cite{kumar2012leafsnap}, food \cite{hou2017vegfru}, trees \cite{wegner2016cataloging}, and cars~\cite{krause20133d,lin2014jointly, yang2015large, gebru2017fine}. ImageNet~\cite{russakovsky2015imagenet} is not typically advertised as a fine-grained dataset, yet contains several groups of fine-grained classes, including about $60$ bird species and about $120$ dog breeds. 
In Table \ref{tab:rel_datasets} we summarize the statistics of some of the most common datasets. 
With the exception of a small number~\eg\cite{krause2016unreasonable, gebru2017fine}, many of these datasets were typically constructed to have an approximately uniform distribution of images across the different categories.
In addition, many of these datasets were created by searching the internet with automated web crawlers and as a result can contain a large proportion of incorrect images \eg \cite{krause2016unreasonable}.
Even manually vetted datasets such as ImageNet \cite{russakovsky2015imagenet} have been reported to contain up to 4\% error for some fine-grained categories \cite{van2015building}. 
While current deep models are robust to label noise at training time, it is still very important to have clean validation and test sets to be able to quantify performance \cite{van2015building, rolnick2017deep}.

Unlike web scraped datasets \cite{krause2016unreasonable, krasin2016openimages, wilber2017bam, hou2017vegfru}, the annotations in iNat2017 represent the consensus of informed enthusiasts. 
Images of natural species tend to be challenging as individuals from the same species can differ in appearance due to sex and age, and may also appear in different environments. 
Depending on the particular species, they can also be very challenging to photograph in the wild.
In contrast, mass-produced, man-made object categories are typically identical up to nuisance factors, \ie they only differ in terms of pose, lighting, color, but not necessarily in their underlying object shape or appearance \cite{yu2014fine, gebru2017fine, iMat2017}.

\begin{table}[t]
\small
\begin{center}
\begin{tabular}{ |l|r|r|r| }\hline 
{\bf Dataset Name} & {\bf\# Train}&{\bf \# Classes}&{\bf Imbalance}\\\hline
Flowers 102 \cite{nilsback2006visual} & 1,020 & 102 & 1.00\\
Aircraft \cite{maji2013fine} & 3,334 & 100 & 1.03 \\ 
Oxford Pets \cite{parkhi12a} & 3,680 & 37 &  1.08\\  
DogSnap \cite{liu2012dog} & 4,776 & 133 & 2.85 \\
CUB 200-2011 \cite{wah2011caltech} & 5,994  & 200  & 1.03  \\
Stanford Cars \cite{krause20133d} & 8,144 & 196 & 2.83 \\
Stanford Dogs \cite{KhoslaYaoJayadevaprakashFeiFei_FGVC2011} & 12,000 & 120 & 1.00 \\   
Urban Trees \cite{wegner2016cataloging} &14,572 & 18 & 7.51\\  
NABirds \cite{van2015building} & 23,929 & 555 & 15.00 \\ 
LeafSnap$^*$ \cite{kumar2012leafsnap} & 30,866 & 185 & 8.00\\
CompCars$^*$ \cite{yang2015large} & 136,727 & 1,716 & 10.15\\   
VegFru$^*$ \cite{hou2017vegfru} & 160,731 & 292 & 8.00\\
Census Cars \cite{gebru2017fine} & 512,765 & 2,675 & 10.00\\ %
ILSVRC2012 \cite{russakovsky2015imagenet} & 1,281,167 & 1,000 & 1.78 \\ 
{\bf iNat2017} & 579,184 & 5,089 & 435.44\\ \hline
\end{tabular}
\end{center}
\caption{Summary of popular general and fine-grained computer vision classification datasets. 
`Imbalance' represents the number of images in the largest class divided by the number of images in the smallest. 
While susceptible to outliers, it gives an indication of the imbalance found in many common datasets. 
$^*$Total number of train, validation, and test images.}
\label{tab:rel_datasets}
\end{table}

\section{Dataset Overview}
\label{sec:dataset_overview}

\begin{table}[t]
\small
\begin{center}
\begin{tabular}{ |>{\centering}m{1.0em}|l|r|r|r|r| }\hline 
& {\bf Super-Class}&{\bf Class}&{\bf Train}&{\bf Val}&{\bf BBoxes}\\\hline
\includegraphics[height=8px]{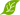}&Plantae&2,101&158,407&38,206&-\\
\includegraphics[height=8px]{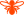}&Insecta&1,021&100,479&18,076&125,679\\
\includegraphics[height=8px]{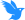}&Aves&964&214,295&21,226&311,669\\
\includegraphics[height=8px]{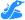}&Reptilia&289&35,201&5,680&42,351\\
\includegraphics[height=8px]{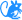}&Mammalia&186&29,333&3,490&35,222\\
\includegraphics[height=8px]{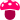}&Fungi&121&5,826&1,780&-\\
\includegraphics[height=8px]{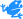}&Amphibia&115&15,318&2,385&18,281\\
\includegraphics[height=8px]{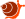}&Mollusca&93&7,536&1,841&10,821\\
\includegraphics[height=8px]{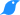}&Animalia&77&5,228&1,362&8,536\\
\includegraphics[height=8px]{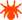}&Arachnida&56&4,873&1,086&5,826\\
\includegraphics[height=8px]{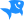}&Actinopterygii&53&1,982&637&3,382\\
\includegraphics[height=8px]{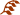}&Chromista&9&398&144&-\\
\includegraphics[height=8px]{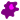}&Protozoa&4&308&73&-\\
\hline
& {\bf Total}&5,089&579,184&95,986&561,767\\\hline
\end{tabular}
\end{center}
\caption{Number of images, classes, and bounding boxes in iNat2017 broken down by super-class.
`Animalia' is a catch-all category that contains species that do not fit in the other super-classes.
Bounding boxes were collected for nine of the super-classes. 
In addition, the public and private test sets contain 90,427 and 92,280 images, respectively.}
\label{tab:data_overview}
\end{table}


In this section we describe the details of the dataset, including how we collected the image data (Section ~\ref{sec:dataset_collection}), how we constructed the train, validation and test splits (Section ~\ref{sec:dataset_splits}), how we vetted the test split (Section ~\ref{sec:dataset_verification}) and how we collected bounding boxes (Section~\ref{sec:dataset_bboxes}). Future researchers may find our experience useful when constructing their own datasets.


\subsection{Dataset Collection}
\label{sec:dataset_collection}
iNat2017 was collected in collaboration with iNaturalist, a citizen science effort that allows naturalists to map and share observations of biodiversity across the globe through a custom made web portal and mobile apps. Observations, submitted by \textit{observers}, consist of images, descriptions, location and time data, and community identifications. If the community reaches a consensus on the taxa in the observation, then a ``research-grade'' label is applied to the observation. 
iNaturalist makes an archive of research-grade observation data available to the environmental science community via the Global Biodiversity Information Facility (GBIF) \cite{iNatGBIF}. Only research-grade labels at genus, species or lower are included in this archive. These archives contain the necessary information to reconstruct which photographs belong to each observation, which observations belong to each observer, as well as the taxonomic hierarchy relating the taxa. These archives are refreshed on a rolling basis and the iNat2017 dataset was created by processing the archive from October 3rd, 2016.

\begin{figure}[t]
\centering
\includegraphics[width=\columnwidth]{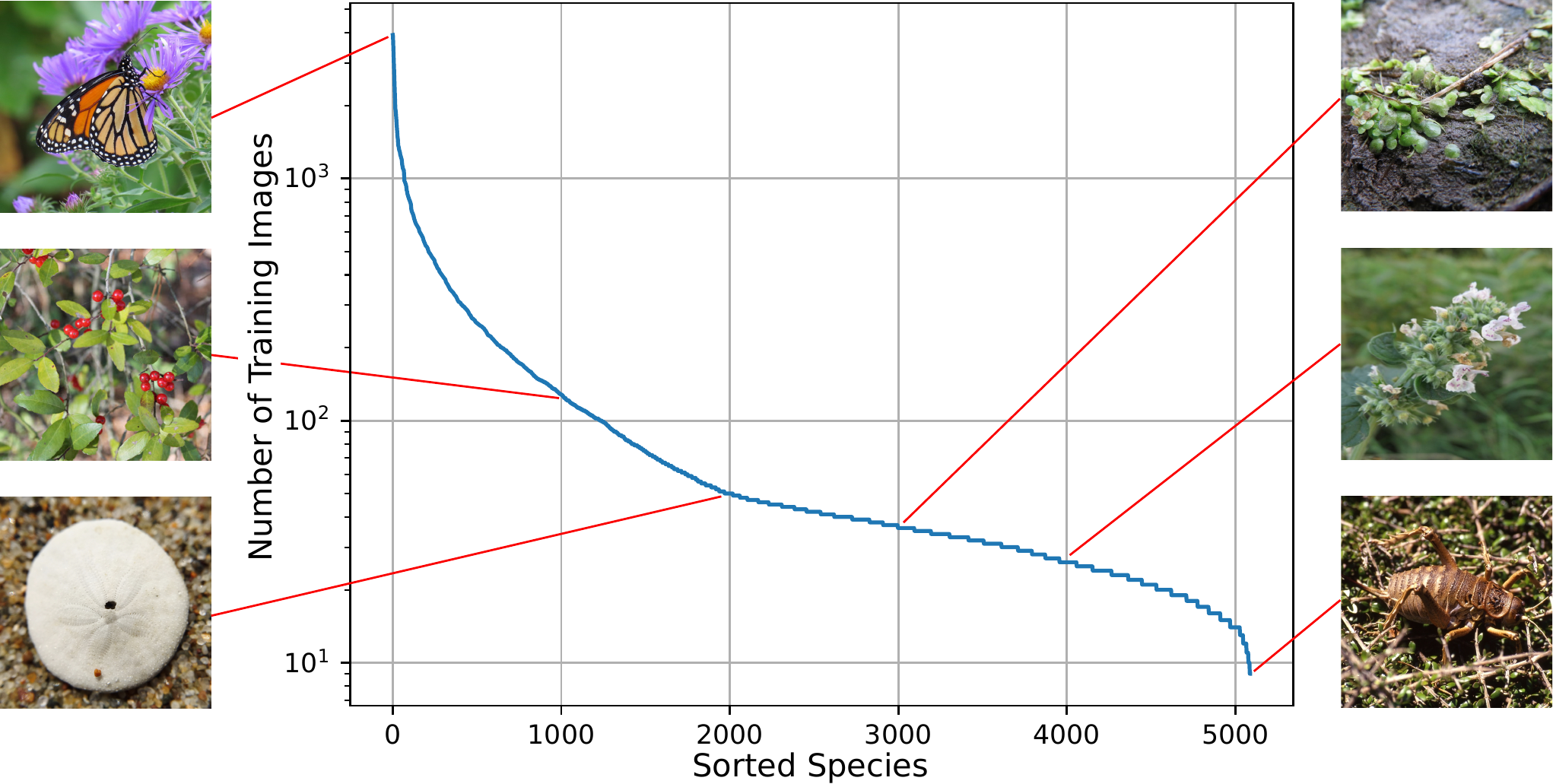}
\caption{Distribution of training images per species. iNat2017 contains a large imbalance between classes, where the top 1\% most populated classes contain over 16\% of training images.} 
\label{fig:num_ims}
\end{figure}

\subsection{Dataset Construction}
\label{sec:dataset_splits}
The complete GBIF archive had 54k classes (genus level taxa and below), with 1.1M observations and a total of 1.6M images. However, over 19k of those classes contained only one observation. In order to construct train, validation  and test splits that contained samples from all classes we chose to employ a taxa selection criteria: we required that a taxa have at least 20 observations, submitted from at least 20 unique observers (i.e. one observation from each of the 20 unique observers). This criteria limited the candidate set to 5,089 taxa coming from 13 super-classes, see Table \ref{tab:data_overview}. 

The next step was to partition the images from these taxa into the train, validation, and test splits. For each of the selected taxa, we sorted the \textit{observers} by their number of observations (fewest first) and selected the first 40\% of observers to be in the test split, and the remaining 60\% to be in the ``train-val'' split. By partitioning the observers in this way, and subsequently placing all of their photographs into one split or the other, we ensure that the behavior of a particular user (\eg camera equipment, location, background, \etc) is contained within a single split, and not available as a useful source of information for classification on the other split for a specific taxa. Note that a particular observer may be put in the test split for one taxa, but the ``train-val'' split for another taxa. By first sorting the observers by their number of observations we ensure that the test split contains a high number of unique observers and therefore a high degree of variability. To be concrete, at this point, for a taxa that has exactly 20 unique observers (the minimum allowed), 8 observers would be placed in the the test split and the remaining 12 observers would be placed in the ``train-val'' split. Rather than release all test images, we randomly sampled $\sim$183,000 to be included in the final dataset. The remaining test images were held in reserve in case we encountered unforeseen problems with the dataset.

To construct the separate train and validation splits for each taxa from the ``train-val'' split we again partition on the observers. For each taxa, we sort the observers by increasing observation counts and repeatedly add observers to the validation split until either of the following conditions occurs: (1) The total number of \textit{photographs} in the validation set exceeds 30, or (2) 33\% of the available \textit{photographs} in the ``train-val'' set for the taxa have been added to the validation set. The remaining observers and all of their photographs are added to the train split. To be concrete, and continuing the example from above, exactly 4 images would be placed in the validation split, and the remaining 8 images would be placed in the train split for a taxa with 20 unique observers. This results in a validation split that has at least 4 and at most $\sim$30 images for each class (the last observer added to the validation split for a taxa may push the number of photographs above 30), and a train split that has at least 8 images for each class. See Fig.~\ref{fig:num_ims} for the distribution of train images per class.

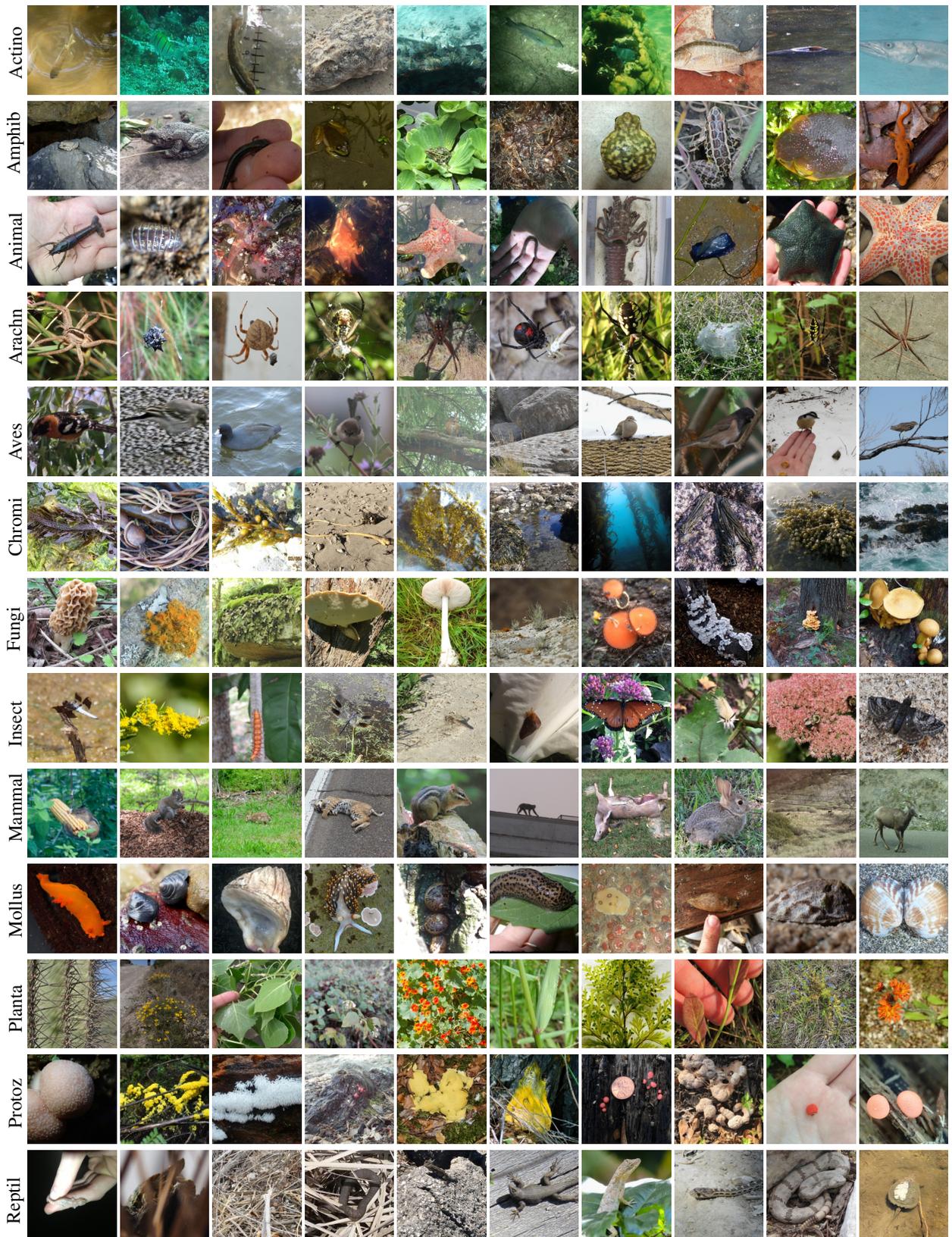
\begin{figure*}
  \centering
  \resizebox{0.98\textwidth}{!}{
      \input{inat_figs/train_ims/train_ims.tex}

  }
  \caption{Example images from the training set. Each row displays randomly selected images from each of the 13 different super-classes. For ease of visualization we show the center crop of each image.}
  \label{fig:train_ims}
\end{figure*}

At this point we have the final image splits, with a total of 579,184 training images, 95,986 validation images and 182,707 test images. 
All images were resized to have a max dimension of 800px. Sample images from the dataset can be viewed in Fig.~\ref{fig:train_ims}.
The iNat2017 dataset is available from our project website\footnote{\url{https://github.com/visipedia/inat_comp/tree/master/2017}}.

\begin{figure}[t]
\centering
\includegraphics[width=\columnwidth]{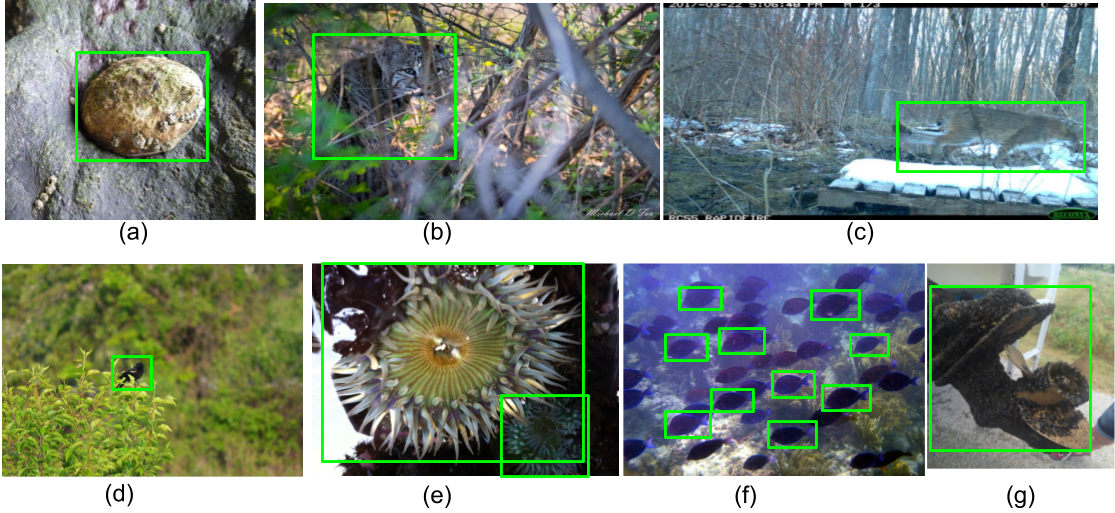}
\caption{Sample bounding box annotations. Annotators were asked to annotate up to 10  instances of a super-class, as opposed to the fine-grained class, in each image.} 
\label{fig:bbox_examples}
\end{figure}

\subsubsection{Test Set Verification}
\label{sec:dataset_verification}
Each observation on iNaturalist is made up of one or more images that provide evidence that the taxa \emph{was present}. Therefore, a small percentage of images may not contain the taxa of interest but instead can include footprints, feces, and habitat shots. Unfortunately, iNaturalist does not distinguish between these types of images in the GBIF export, so we crowdsourced the verification of three super-classes (Mammalia, Aves, and Reptilia) that might exhibit these ``non-instance'' images. We found that less than 1.1\%  of the test set images for Aves and Reptilia had non-instance images. The fraction was higher for Mammalia due to the prevalence of footprint and feces images, and we filtered these images out of the test set. The training and validation images were not filtered.

\begin{figure}[t]
\centering
\includegraphics[width=\columnwidth]{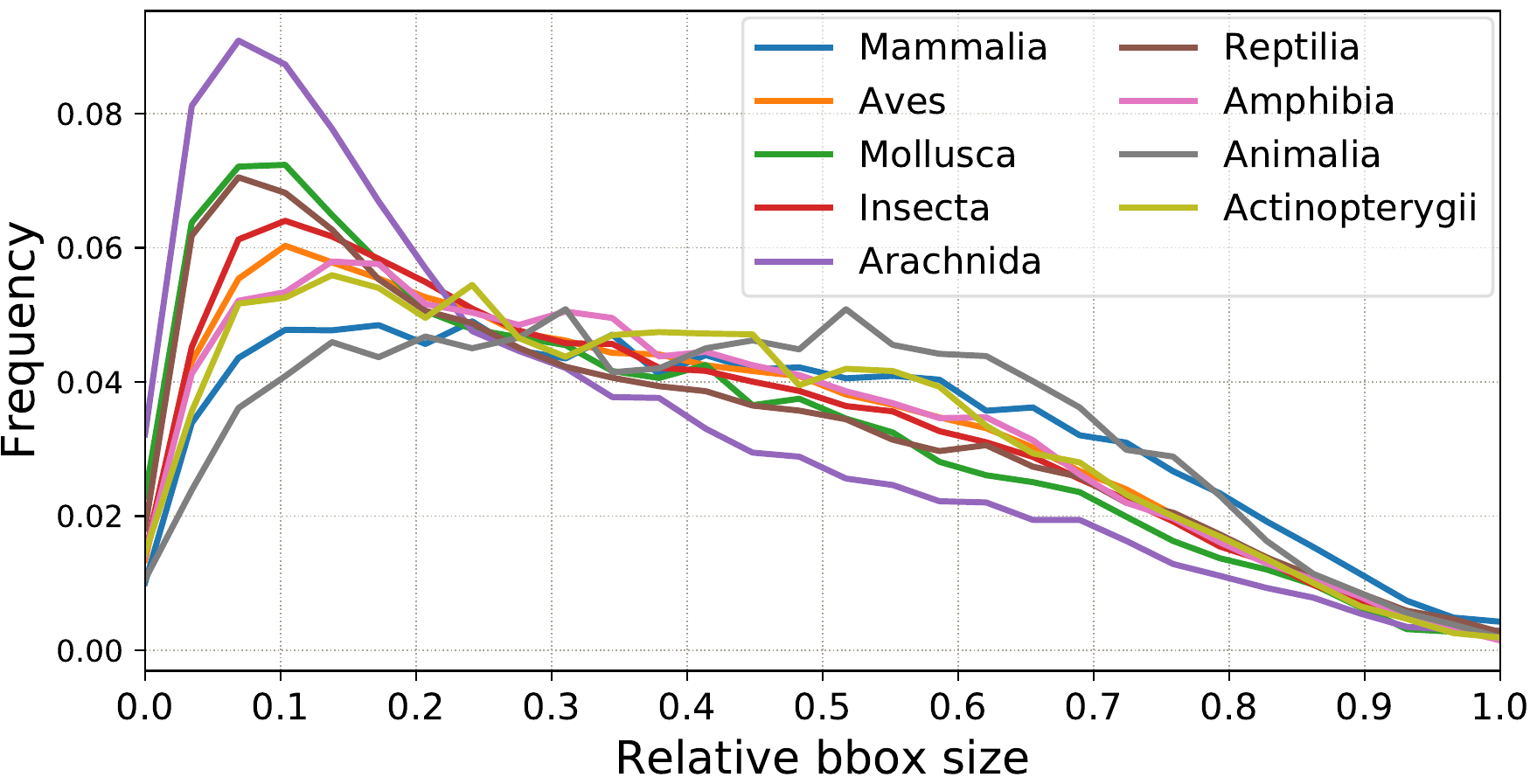}
\caption{The distribution of relative bounding box sizes (calculated by $\sqrt{w_{bbox} \times h_{bbox}} / \sqrt{w_{img} \times h_{img}}$) in the training set, per super-class. Most objects are relatively small or medium sized.
} 
\label{fig:bbox_size}
\vspace{-3mm}
\end{figure}

\subsection{Bounding Box Annotation}
\label{sec:dataset_bboxes}
Bounding boxes were collected on 9 out of the 13 super-classes (see Table~\ref{tab:data_overview}), totaling 2,854 classes. Due to the inherit difficultly of asking non-expert crowd annotators to both recognize and box specific fine-grained classes, we instructed annotators to instead box all instances of the associated super-class for a taxa (\eg ``Box all Birds'' rather than ``Box all Red-winged Black Birds''). We collected super-class boxes only on taxa that are part of that super-class. For some super-classes (\eg Mollusca), there are images containing taxa which are unfamiliar to many of the annotators (\eg Fig.~\ref{fig:bbox_examples}(a)). For those cases, we instructed the annotators to box the prominent objects in the images.  

The task instructions specified to draw boxes tightly around all parts of the animal (including legs, horns, antennas, \etc). If the animal is occluded, the annotators were instructed to draw the box around the visible parts (\eg Fig.~\ref{fig:bbox_examples}(b)). In cases where the animal is blurry or small (\eg Fig.~\ref{fig:bbox_examples}(c) and (d)), the following rule-of-thumb was used:  ``if you are confident that it is an animal from the requested super-class, regardless of size, blurriness or occlusion, put a box around it.'' For images with multiple instances of the super-class, all of them are boxed, up to a limit of 10 (Fig.~\ref{fig:bbox_examples}(f)), and bounding boxes may overlap (Fig.~\ref{fig:bbox_examples}(e)). We observe that $12\%$ of images have more than 1 instance and $1.3\%$ have more than 5. If the instances are physically connected (\eg the mussels in Fig.~\ref{fig:bbox_examples}(g)), then only one box is placed around them. 

Bounding boxes were not collected on the Plantae, Fungi, Protozoa or Chromista super-classes because these super-classes exhibit properties that make it difficult to box the individual instances (\eg close up of trees, bushes, kelp, \etc). An alternate form of pixel annotations, potentially from a more specialized group of crowd workers, may be more appropriate for these classes. 

Under the above guidelines, 561,767 bounding boxes were obtained from 449,313 images in the training and validation sets. Following the size conventions of COCO~\cite{lin2014microsoft}, the iNat2017 dataset is composed of $5.7\%$ small instances (area $< 32^2$), $23.6\%$ medium instances ($32^2 \leq$ area $\leq 96^2$) and $70.7\%$  large instances (area $> 96^2$), with area computed as $50\%$ of the annotated bounding box area (since segmentation masks were not collected). Fig.~\ref{fig:bbox_size} shows the distribution of relative bounding box sizes, indicating that a majority of instances are relatively small and medium sized.

\section{Experiments}
In this section we compare the performance of state-of-the-art classification and detection models on iNat2017.

\subsection{Classification Results}
To characterize the classification difficulty of iNat2017, we ran experiments with several state-of-the-art deep network architectures, including ResNets \cite{he2016deep}, Inception V3~\cite{szegedy2016rethinking}, Inception ResNet V2~\cite{szegedy2016inception} and MobileNet \cite{howard2017mobilenets}.
During training, random cropping with aspect ratio augmentation \cite{googlenet} was used. 
Training batches of size 32 were created by uniformly sampling from all available training images as opposed to sampling uniformly from the classes. 
We fine-tuned all networks from ImageNet pre-trained weights with a learning rate of 0.0045, decayed exponentially by 0.94 every 4 epochs, and RMSProp optimization with momentum and decay both set to 0.9.
Training and testing were performed with an image size of $299 \times 299$, with a single centered crop at test time.

Table \ref{tab:class_baselines} summarizes the top-1 and top-5 accuracy of the models. From the Inception family, we see that the higher capacity Inception ResNet V2 outperforms the Inception V3 network. The addition of the Squeeze-and-Excitation (SE) blocks \cite{hu2017squeeze} further improves performance for both models by a small amount.
ResNets performed worse on iNat2017 compared to the Inception architectures, likely due to over-fitting on categories with small number of training images.
We found that adding a 0.5 probability dropout layer (drp) could improve the performance of ResNets. 
MobileNet, designed to efficiently run on embedded devices, had the lowest performance.

Overall, the Inception ResNetV2 SE was the best performing model. As a comparison, this model achieves a single crop top-1 and top-5 accuracy of 80.2\% and 95.21\% respectively on the ILSVRC 2012 \cite{russakovsky2015imagenet} validation set \cite{szegedy2016inception}, as opposed to 67.74\% and 87.89\% on iNat2017, highlighting the comparative difficulty of the iNat2017 dataset. A more detailed super-class level breakdown is available in Table \ref{tab:sc_acc} for the Inception ResNetV2 SE model. We can see that the Reptilia super-class (with 289 classes) was the most difficult with an average top-1 accuracy of 45.87\%, while the Protozoa super-class (with 4 classes) had the highest accuracy at 89.19\%. Viewed as a collection of fine-grained datasets (one for each super-class) we can see that the iNat2017 dataset exhibits highly variable classification difficulty. 

\begin{figure}[h!]
\centering
\includegraphics[width=\columnwidth]{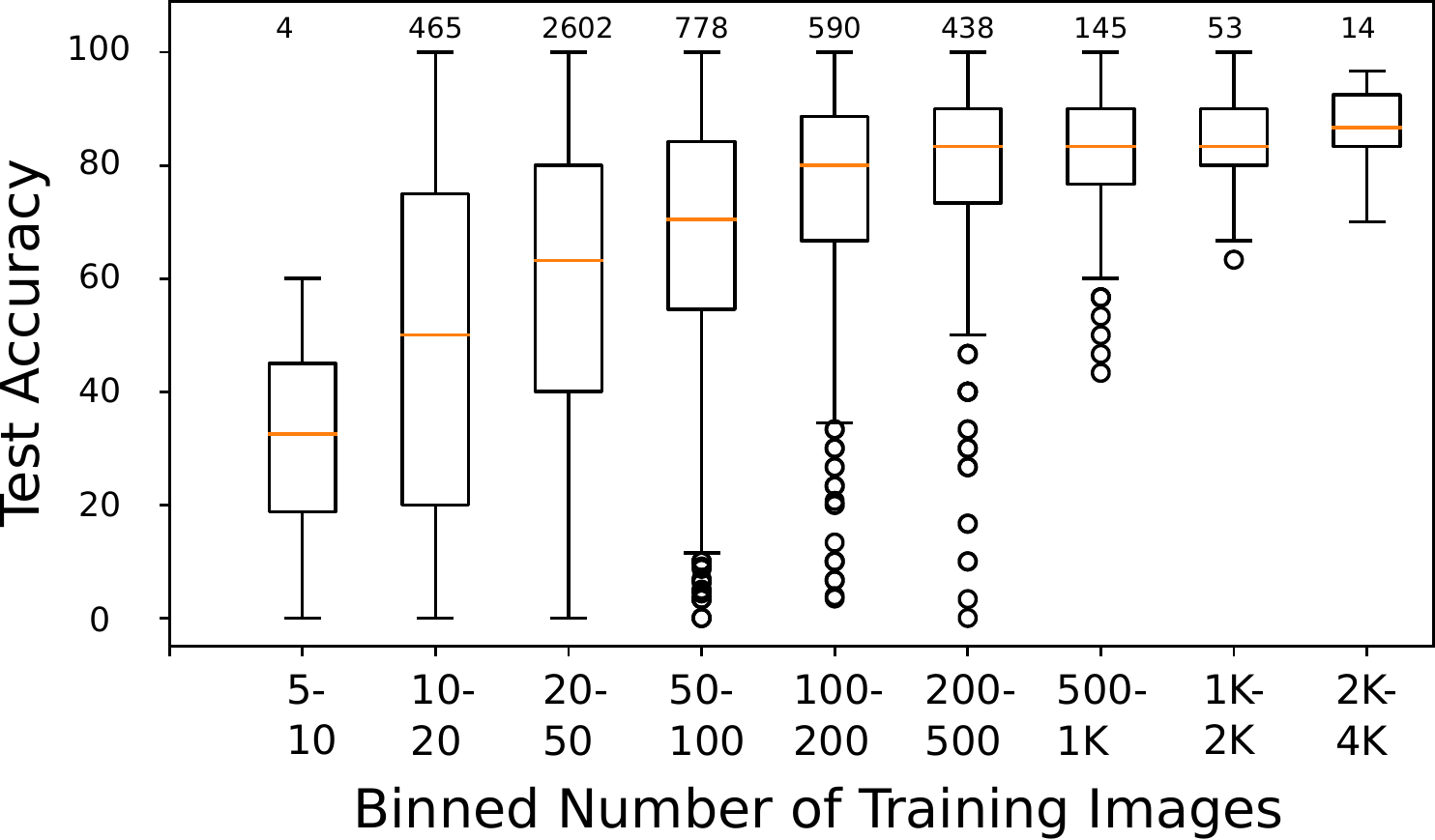}
\caption{Top one public test set accuracy per class for IncResNet V2 SE \cite{hu2017squeeze}.
Each box plot represents classes grouped by the number of training images. 
The number of classes for each bin is written on top of each box plot. Performance improves with the number of training images, but the challenge is how to maintain high accuracy with fewer images?
} 
\label{fig:acc_vs_train}
\vspace{-2mm}
\end{figure}

In Fig.~\ref{fig:acc_vs_train} we plot the top one public test set accuracy against the number of training images for each class from the Inception ResNet V2 SE model. 
We see that as the number of training images per class increases, so does the test accuracy. 
However, we still observe a large variance in accuracy for classes with a similar amount of training data, revealing opportunities for algorithmic improvements in both the low data and high data regimes.

\begin{table}[t]
\footnotesize
\begin{center}
\begin{tabular}{ l|r|r|r|r|r|r } 
 \hline
 & \multicolumn{2}{|c}{\bf Validation} & \multicolumn{2}{|c}{\bf Public Test} & \multicolumn{2}{|c}{\bf Private Test} \\ \hline
 & Top1 & Top5 & Top1 & Top5      & Top1 & Top5 \\ \hline
 IncResNetV2 SE        & {\bf 67.3} & {\bf 87.5} & {\bf 68.5}  & {\bf 88.2}         & 67.7 & {\bf 87.9}\\ 
 IncResNetV2 & 67.1 & {\bf 87.5} & 68.3 &	88.0 & {\bf 67.8} &	87.8\\ 
 IncV3 SE               & 65.0 & 85.9 & 66.3 &	86.7        & 65.2 &	86.3\\ 
 IncV3 & 64.2 & 85.2 & 65.5 &	86.1 & 64.8 &	85.7\\ 
 ResNet152 drp           & 62.6 & 84.5 & 64.2 &	85.5    & 63.1 &	85.1\\ 
 ResNet101 drp          & 60.9 & 83.1 & 62.4 &	84.1     & 61.4 &	83.6\\ 
 ResNet152               & 59.0 & 80.5    & 60.6 & 	81.7     & 59.7 &	81.3\\  
 ResNet101              & 58.4 & 80.0     & 59.9 &	81.2     & 59.1 &	80.9\\  
 MobileNet V1          & 52.9 & 75.4   & 54.4 &	76.8 & 53.7 &	76.3\\ \hline
\end{tabular}
\end{center}
\caption{Classification results for various CNNs trained on only the training set, using a single center crop at test time. Unlike some current datasets where performance is near saturation, iNat2017 still poses a challenge for state-of-the-art classifiers.}
\label{tab:class_baselines}
\end{table}

\begin{table}[t]
\footnotesize
\begin{center}
\begin{tabular}{ |l|r|r|r| }\hline 
{\bf Super-Class} & {\bf Avg Train} & \multicolumn{2}{c|}{\bf Public Test} \\ \hline
& & Top1 &Top5 \\\hline
Plantae    & 75.4 &	69.5 & 87.1 \\
Insecta    & 98.4 &	77.1 & 93.4\\
Aves       & 222.3&	67.3 & 88.0\\
Reptilia   & 121.8& 45.9 & 80.9\\
Mammalia   & 157.7& 61.4  & 85.1 \\
Fungi      & 48.1& 	74.0  & 92.3 \\
Amphibia   & 67.9& 	51.2  & 81.0 \\
Mollusca   & 81.0& 	72.4  & 90.9\\
Animalia   & 67.9 &	73.8  & 91.1\\
Arachnida  & 87.0 &	71.5 & 88.8\\
Actinopterygii & 37.4& 	70.8  & 86.3\\
Chromista      & 44.2& 	73.8  & 92.4\\
Protozoa       & 77.0& 	89.2 &  96.0 \\ \hline
\end{tabular}
\end{center}
\caption{Super-class level accuracy (computed by averaging across all species within each super-class) for the best performing model Inception ResNetV2 SE \cite{hu2017squeeze}. 
``Avg Train'' indicates the average number of training images per class for each super-class. 
We observe a large difference in performance across the different super-classes.
}
\label{tab:sc_acc}
\end{table}

\vspace{-10mm}

\subsection{Detection Results}
To characterize the detection difficulty of iNat2017, we adopt Faster-RCNN~\cite{ren2017faster} for its state-of-the-art performance as an object detection setup (which jointly predicts object bounding boxes along with class labels). We use a TensorFlow~\cite{abadi2016tensorflow} implementation of Faster-RCNN with default hyper-parameters~\cite{huang2017speed}. Each model is trained with 0.9 momentum, and asynchronously optimized on 9 GPUs to expedite experiments. We use an Inception V3 network, initialized from ImageNet, as the backbone for our Faster-RCNN models. Finally, each input image is resized to have 600 pixels as the short edge while maintaining the aspect ratio.

As discussed in Section~\ref{sec:dataset_bboxes}, we collected bounding boxes on 9 of the 13 super-classes, translating to a total of 2,854 classes with bounding boxes. In the following experiments we only consider performance on this subset of classes. Additionally, we report performance on the the validation set in place of the test set and we only evaluate on images that contained a single instance. Images that contained only evidence of the species' presence and images that contained multiple instances were excluded. We evaluate the models using the detection metrics from COCO~\cite{lin2014microsoft}.

\begin{figure}[t!]
\centering
\includegraphics[width=0.95\columnwidth]{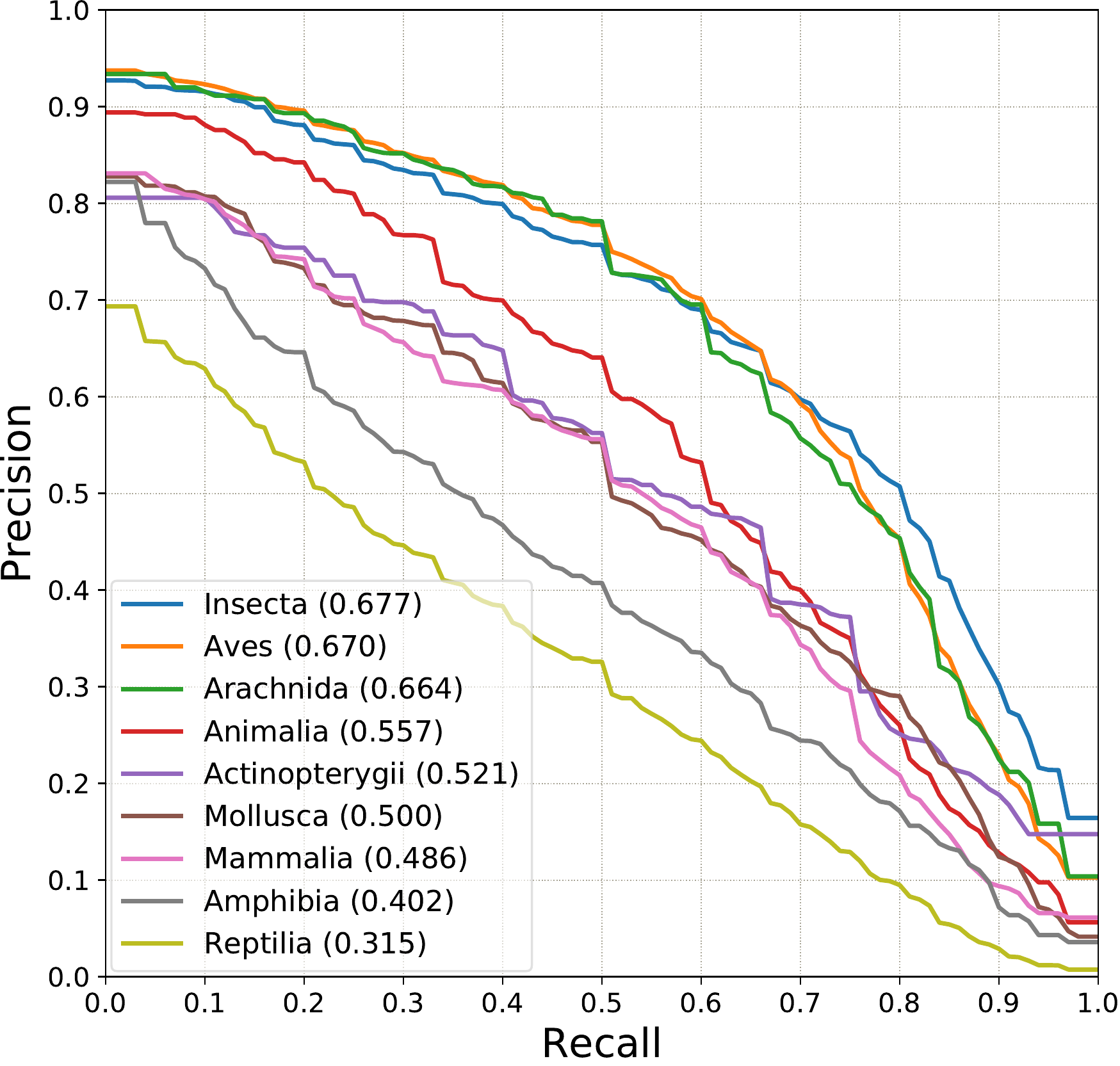}
\caption{Precision-Recall curve with 0.5 IoU for each super-class, where the Area-Under-Curve (AUC) corresponds to AP$^{50}$ in Table~\ref{tab:ap_ar}. Super-class performance is calculated by averaging across all fine-grained classes. We can see that building a detector that works well for all super-classes in iNat2017 will be a challenge. }
\label{fig:pr_curve}
\end{figure}

\begin{figure*}[h!]
\centering
\includegraphics[width=\textwidth]{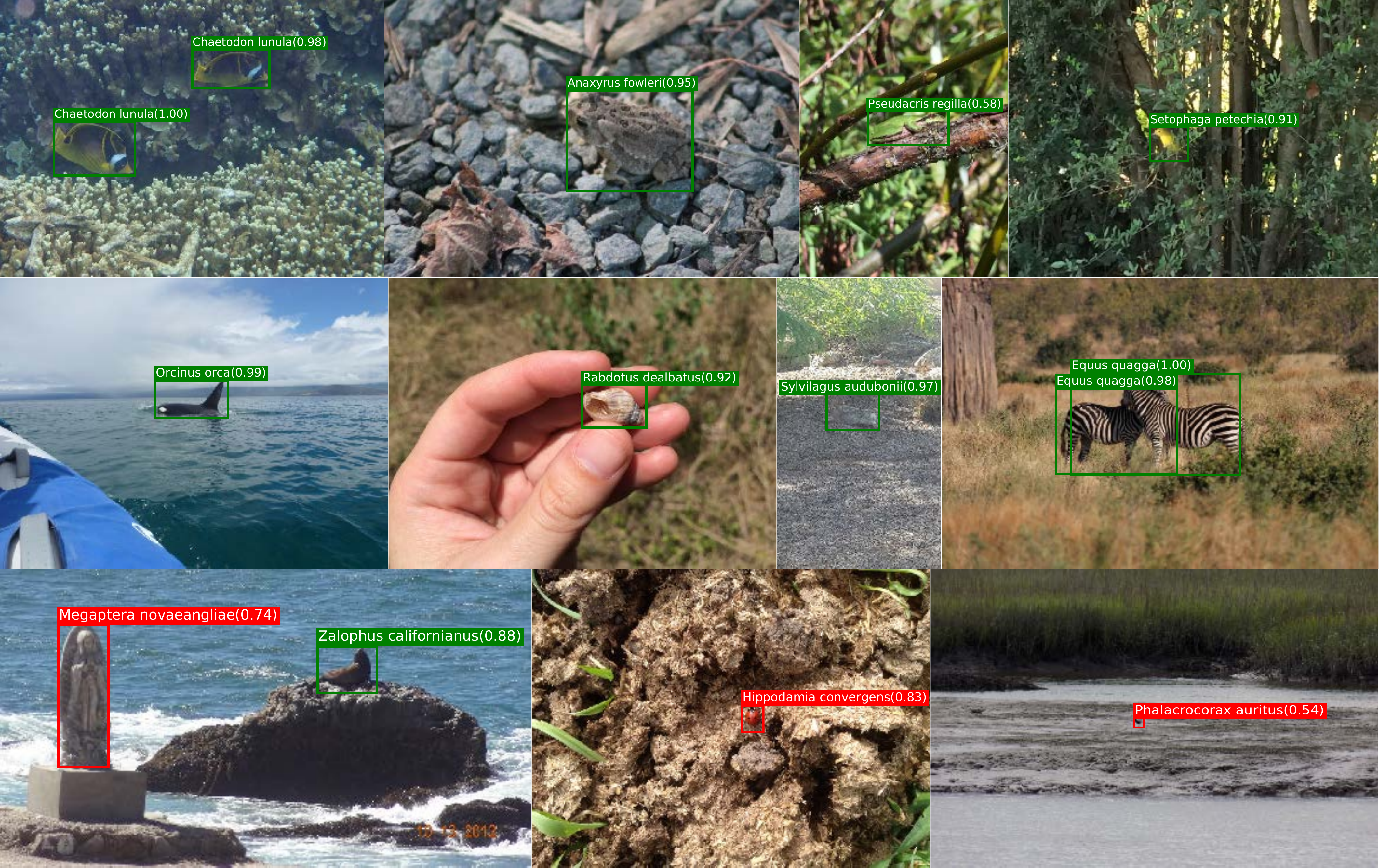}
\caption{Sample detection results for the 2,854-class model that was evaluated across all validation images. Green boxes represent correct species level detections, while reds are mistakes. 
The bottom row depicts some failure cases. We see that small objects pose a challenge for classification, even when localized well.} 
\label{fig:detection_results}
\end{figure*}

We first study the performance of fine-grained localization and classification by training the Faster-RCNN model on the 2,854 class subset. Fig.~\ref{fig:detection_results} shows some sample detection results. Table~\ref{tab:ap_ar} provides the break down in performance for each super-class, where super-class performance is computed by taking an average across all classes within the super-class. The precision-recall curves (again at the super-class level) for 0.5 IoU are displayed in Fig.~\ref{fig:pr_curve}. Across all super-classes we achieve a comprehensive average precision (AP) of 43.5. Again the Reptilia super-class proved to be the most difficult, with an AP of 21.3 and an AUC of 0.315. At the other end of the spectrum we achieved an AP of 49.4 for Insecta and an AUC of 0.677. Similar to the classification results, when viewed as a a collection of datasets (one for each super-class) we see that iNat2017 exhibits highly variable detection difficulty, posing a challenge to researchers to build improved detectors that work across a broad group of fine-grained classes.

Next we explored the effect of label granularity on detection performance. We trained two more Faster-RCNN models, one trained to detect super classes rather fine-grained classes (so 9 classes in total) and another model trained with all labels pooled together, resulting in a generic object / not object detector. Table~\ref{tab:coco_ap} shows the resulting AP scores for the three models when evaluated at different granularities. When evaluated on the coarser granularity, detectors trained on finer-grained categories have lower detection performance when compared with detectors trained at coarser labels. The performance of the 2,854-class detector is particularly poor on super-class recognition and object localization. This suggests that the Faster-RCNN algorithm has plenty of room for improvements on end-to-end fine-grained detection tasks.

\begin{table}[t]
\footnotesize
\begin{center}
\begin{tabular}{ l|r|r|r|r|r } 
 \hline
 & \textbf{AP} & \textbf{AP$^{50}$} & \textbf{AP$^{75}$} & \textbf{AR$^{1}$} & \textbf{AR$^{10}$} \\ \hline
 Insecta & 49.4 & \textbf{67.7} & \textbf{59.3} & \textbf{64.5} & \textbf{64.9} \\
 Aves & \textbf{49.5} & 67.0 & 59.1 & 63.3 & 63.6 \\
 Reptilia & 21.3 & 31.5 & 24.9 & 44.0 & 44.8 \\
 Mammalia & 33.3 & 48.6 & 39.1 & 49.8 & 50.6 \\
 Amphibia & 28.7 & 40.2 & 35.0 & 52.0 & 52.3 \\
 Mollusca & 34.8 & 50.0 & 41.6 & 52.0 & 53.0 \\
 Animalia & 35.6 & 55.7 & 40.8 & 48.3 & 50.5 \\
 Arachnida & 43.9 & 66.4 & 49.6 & 57.3 & 58.6 \\
 Actinopterygii & 35.0 & 52.1 & 41.6 & 49.1 & 49.6 \\ \hline
 Overall & 43.5 & 60.2 & 51.8 & 59.3 & 59.8 \\ \hline
\end{tabular}
\end{center}
\caption{Super-class-level Average Precision (AP) and Average Recall (AR) for object detection, where AP, AP$^{50}$ and AP$^{75}$ denotes AP@[IoU=.50:.05:.95], AP@[IoU=.50] and AP@[IoU=.75] respectively; AR$^{1}$ and AR$^{10}$ denotes AR given 1 detection and 10 detections per image.}
\label{tab:ap_ar}
\end{table}

\begin{table}[t]
\small
\begin{center}
\begin{tabular}{ |l|c|c|c| } 
 \hline
 {\bf Training} & \multicolumn{3}{c|}{\bf Evaluation} \\ \hline
                  & 2854-class & 9-super-class & 1-generic \\ \hline
 2854-class       & 43.5    & 55.6 & 63.7 \\ 
 9-super-class    & -       & 65.8 & 76.7 \\ 
 1-generic        & -       & -    & 78.5 \\ \hline
\end{tabular}
\end{center}
\caption{Detection performance (AP@[IoU=.50:.05:.95]) with different training and evaluation class granularity. Using finer-grained class labels during training has a negative impact on coarser-grained super-class detection. This presents an opportunity for new detection algorithms that maintain precision at the fine-grained level.}
\label{tab:coco_ap}
\end{table}

%
%

\section{Conclusions and Future Work}
We present the iNat2017 dataset, in contrast to many existing computer vision datasets it is: 1) unbiased, in that it was collected by non-computer vision researchers for a well defined purpose, 2) more representative of real-world challenges than previous datasets, 3) represents a long-tail classification problem, and 4) is useful in conservation and field biology.
The introduction of iNat2017 enables us to study two important questions in a real world setting: 1) do long-tailed datasets present intrinsic challenges? and 2) do our computer vision systems exhibit transfer learning from the well-represented categories to the least represented ones?
While our baseline classification and detection results are encouraging, from our experiments we see that state-of-the-art computer vision models have room to improve when applied to large imbalanced datasets. Small efficient models designed for mobile applications and embedded devices have even more room for improvement~\cite{howard2017mobilenets}.

Unlike traditional, researcher-collected datasets, the iNat2017 dataset has the opportunity to grow with the iNaturalist community. Currently, every 1.7 hours another species passes the 20 unique observer threshold, making it available for inclusion in the dataset (already up to 12k as of November 2017, up from 5k when we started work on the dataset). Thus, the current challenges of the dataset (long tail with sparse data) will only become more relevant. 

In the future we plan to investigate additional annotations such as sex and life stage attributes, habitat tags, and pixel level labels for the four super-classes that were challenging to annotate. We also plan to explore the ``open-world problem'' where the test set contains classes that were never seen during training. This direction would encourage new error measures that incorporate taxonomic rank~\cite{mittal2012taxonomic, yan2015hd}. Finally, we expect this dataset to be useful in studying how to teach fine-grained visual categories to humans \cite{singla2014near, johns2015becoming}, and plan to experiment with models of human learning. 


\vspace{5pt}
\par\noindent\textbf{Acknowledgments.} 
This work was supported by a Google Focused Research Award. We would like to thank: Scott Loarie and Ken-ichi Ueda from iNaturalist; Steve Branson, David Rolnick, Weijun Wang, and Nathan Frey for their help with the dataset; Wendy Kan and Maggie Demkin from Kaggle; the iNat2017 competitors, and the FGVC2017 workshop organizers. We also thank NVIDIA and Amazon Web Services for their donations.


{\small
\bibliographystyle{ieee}
\bibliography{iNat}
}

\end{document}


\title{The iNaturalist Species Classification and Detection Dataset\\- Supplementary Material}

\author{Grant Van Horn$^{1}$\hspace{15pt}Oisin Mac Aodha$^{1}$\hspace{15pt}Yang Song$^{2}$\hspace{15pt}Yin Cui$^{3}$\hspace{15pt}Chen Sun$^{2}$\\Alex Shepard$^{4}$\hspace{15pt}Hartwig Adam$^{2}$\hspace{15pt}Pietro Perona$^{1}$\hspace{15pt}Serge Belongie$^{3}$\\
\\
$^{1}$Caltech\hspace{20pt}$^{2}$Google\hspace{20pt}$^{3}$Cornell Tech\hspace{20pt}$^{4}$iNaturalist}

\maketitle
\thispagestyle{empty}

\section{Additional Classification Results}

We performed an experiment to understand if there was any relationship between real world animal size and prediction accuracy.
Using existing records for bird \cite{lislevand2007avian} and mammal \cite{jones2009pantheria} body sizes we assigned a mass to each of the classes in iNat2017 that overlapped with these datasets. 
For a given species, mass will vary due to the life stage or gender of the particular individual.
Here, we simply take the average value. 
This resulted in data for 795 species, from the small Allen's hummingbird ({\it Selasphorus sasin}) to the large Humpback whale {\it Megaptera novaeangliae}. 
In Fig.~\ref{fig:acc_vs_mass} we can see that median accuracy decreases as the mass of the species increases. 
These results are preliminary, but reinforce the observation that it can be challenging for humans to take good photographs of larger mammals. 
More analysis of these failure cases may allow us to produce better, species-specific, instructions for the photographers on iNaturalist.

\begin{figure}[h]
\centering
\includegraphics[width=0.4\textwidth]{inat_figs/mass_resnet.pdf}
\caption{Top one public test set accuracy per class for \cite{szegedy2016inception} for a subset of 795 classes of birds and mammals binned according to mass. 
The number of classes appears to the bottom right of each box.} 
\label{fig:acc_vs_mass}
\end{figure}

The IUCN Red List of Vulnerable Species monitors and evaluates the extinction risk of thousands of species and subspecies \cite{baillie20042004}.
In Fig. \ref{fig:redlist} we plot the Red List status of 1,568 species from the iNat2017 dataset. 
We see that the vast majority of the species are in the `Least Concern' category and that test accuracy decreases as the threatened status increases. This can perhaps be explained by the reduced number of images for these species in the dataset. 

\begin{figure}[t]
\centering
\includegraphics[width=0.4\textwidth]{inat_figs/resnet_redlist.pdf}
\caption{Top one public test set accuracy for \cite{szegedy2016inception} for a subset of 1,568 species binned according to their IUCN Red List of Threatened Species status \cite{baillie20042004}. 
The number of classes appears to the bottom right of each box. } 
\label{fig:redlist}
\end{figure}

Finally, in Fig. \ref{fig:acc_vs_num_train} we examine the relationship between the number of images and the validation accuracy.
The median number of training images per class for our entire training set is 41. 
For this experiment, we capped the maximum number of training images per class to 10, 20, 50, or all, and trained a separate Inception V3 for each case.
This corresponds to starting with 50,000 for the case of 10 images per class  and then doubling the total amount of training data each time. 
For each species, we randomly selected the images up until the maximum amount.
As noted in the main paper, more attention is needed to improve performance in the low data regime.

\begin{figure}[h]
\centering
\includegraphics[width=0.48\textwidth]{inat_figs/vary_train_v3_crop.pdf}
\caption{As the maximum number of training images per class increases so does the accuracy. However, we observe diminishing returns as the number of images increases. Results are plotted on the validation set for the Inception V3 network \cite{szegedy2016rethinking}.} 
\label{fig:acc_vs_num_train}
\vspace{-2mm}
\end{figure}

\subsection{iNat2017 Competition Results}
From April to mid July 2017, we ran a public challenge on the machine learning competition platform Kaggle\footnote{www.kaggle.com/c/inaturalist-challenge-at-fgvc-2017} using iNat2017. 
Similar to the classification tasks in \cite{russakovsky2015imagenet}, we used the top five accuracy metric to rank competitors. We used this metric as some species can only be disambiguated with additional data provided by the observer, such as location or date. 
Additionally, in a small number of cases multiple species may appear in the same image (\eg a bee on a flower). 
Overall, there were 32 submissions and we display the final results for the top five teams in Table \ref{tab:comp_res}.

The top performing entry from \emph{GMV} consisted of an ensemble of Inception V4 and Inception ResNet V2 networks \cite{szegedy2016inception}. 
Each model was first initialized on the ImageNet-1K dataset and then finetuned with the iNat2017 training set along with 90\% of the validation set, utilizing data augmentation at training time. 
The remaining 10\% of the validation set was used for evaluation.  
To compensate for the imbalanced training data, the models were further fine-tuned on the 90\% subset of the validation data that has a more balanced distribution.
To address small object size in the dataset, inference was performed on $560 \times 560$ resolution images using twelve crops per image at test time. 

The additional training data amounts to 15\% of the original training set, which along with the ensembling, multiple test crops, and higher resolution account for the improved 81.58\% top 1 public accuracy compared to our best performing single model which achieved 68.53\%. 


\begin{table}[h]
\footnotesize
\begin{center}
\begin{tabular}{ l|l|r|r|r|r } 
 \hline
  {\bf Rank} & {\bf Team name} & \multicolumn{2}{|c}{\bf Public Test} & \multicolumn{2}{|c}{\bf Private Test} \\ \hline
  &  & Top1 & Top5             & Top1 & Top5 \\ \hline
 1 & GMV        & {\bf 81.58} & {\bf 95.19}          & {\bf 81.28} & {\bf 95.13}\\ 
 2 &  Terry     & 77.18 & 93.60        & 76.76 & 93.50\\ 
 3 & Not hotdog & 77.04 & 93.13    & 76.56 & 93.01\\ 
 4 &  UncleCat  & 77.64 & 93.06     & 77.44 & 92.97\\ 
 5 & DLUT\_VLG  & 76.75 & 93.04     & 76.19 & 92.96\\  \hline
\end{tabular}
\end{center}
\caption{Final public challenge leaderboard results. `Rank' indicates the final position of the team out of 32  competitors. These results are typically ensemble models, trained with higher input resolution, with the validation set as additional training data.}
\label{tab:comp_res}
\end{table}

\section{Additional Detection Results}
In Table~\ref{tab:ap_ar_scale} we investigate detector performance for the 2,854-class model across different bounding box sizes using the size conventions of the COCO dataset~\cite{lin2014microsoft}.
As expected, performance is directly correlated with size, where smaller objects are more difficult to detect. However, examining Table~\ref{tab:val_bbox_sizes} we can see that total number of these small instances is low for most super-classes.

\begin{table}[h]
\footnotesize
\begin{center}
\begin{tabular}{ l|r|r|r|r|r|r } 
 \hline
 & \textbf{AP$^{\textbf{S}}$} & \textbf{AP$^{\textbf{M}}$} & \textbf{AP$^{\textbf{L}}$} & \textbf{AR$^{\textbf{S}}$} & \textbf{AR$^{\textbf{M}}$} & \textbf{AR$^{\textbf{L}}$} \\ \hline
 Insecta & 13.4 & 34.7 & 51.8 & 13.5 & 38.9 & 67.7 \\
 Aves & 11.5 & \cellcolor{green!25}41.7 & \cellcolor{green!25}55.1 & 13.3 & \cellcolor{green!25}49.2 & \cellcolor{green!25}69.9 \\
 Reptilia & \cellcolor{red!25}0.0 & \cellcolor{red!25}12.4 & \cellcolor{red!25}22.0 & \cellcolor{red!25}0.0 & \cellcolor{red!25}16.3 & \cellcolor{red!25}46.5 \\
 Mammalia & 6.7 & 27.8 & 37.1 & 9.0 & 36.1 & 55.8 \\
 Amphibia & \cellcolor{red!25}0.0 & 23.2 & 29.9 & \cellcolor{red!25}0.0 & 28.7 & 54.9 \\
 Mollusca & 17.5 & 30.8 & 35.8 & 17.5 & 33.6 & 55.9 \\
 Animalia & \cellcolor{green!25}24.0 & 22.7 & 37.1 & \cellcolor{green!25}26.7 & 28.2 & 52.0 \\
 Arachnida & 16.2 & 32.9 & 46.5 & 16.2 & 38.5 & 61.6 \\
 Actinopterygii & 5.0 & 16.3 & 36.1 & 5.0 & 17.9 & 51.1 \\ \hline
 Overall & 11.0 & 34.7 & 46.7 & 12.5 & 40.7 & 63.7 \\ \hline
\end{tabular}
\end{center}
\caption{Super-class level Average Precision (AP) and Average Recall (AR) with respect to object sizes. S, M and, L denote small (area $< 32^2$), medium ($32^2 \leq$ area $\leq 96^2$) and, large (area $> 96^2$) objects. The AP for each super-class is calculated by averaging the results for all species belonging to it. Best and worst performance for each metric are marked by green and red, respectively.}
\label{tab:ap_ar_scale}
\end{table}

\begin{table}[h]
\footnotesize
\begin{center}
\begin{tabular}{ l|r|r|r}
\hline
 & \textbf{Small} & \textbf{Medium} & \textbf{Large} \\ \hline
Insecta & 445 & 2432 & 16429 \\
Aves & 2375 & 8898 & 16239 \\
Reptilia & 32 & 400 & 5426 \\
Mammalia & 280 & 1068 & 2751 \\
Amphibia & 20 & 253 & 2172 \\
Mollusca & 74 & 466 & 1709 \\
Animalia & 72 & 414 & 1404 \\
Arachnida & 12 & 152 & 909 \\
Actinopterygii & 32 & 144 & 634 \\ \hline
\end{tabular}
\end{center}
\caption{The number of super-class instances at each bounding box size in the validation set. While AP and AR is low for some super-classes at a particular size (see Table~\ref{tab:ap_ar_scale}), the actual number of instances at that size may also be low. }
\label{tab:val_bbox_sizes}
\end{table}


%
%
{\small
\bibliographystyle{ieee}
\bibliography{iNat}
}

%% file: inat_figs/train_ims/train_ims.tex
\newcommand{\turnheightnew}{0.195\columnwidth}

\centering
\begin{tabular}{c@{\hskip 2pt}c@{\hskip 2pt}c@{\hskip 2pt}c@{\hskip 2pt}c@{\hskip 2pt}c@{\hskip 2pt}c@{\hskip 2pt}c@{\hskip 2pt}c@{\hskip 2pt}c@{\hskip 2pt}c@{\hskip 2pt}c@{}}
		{\rotatebox{90}{\hspace{8pt}Actino}} 
		 & \includegraphics[height=\turnheightnew]{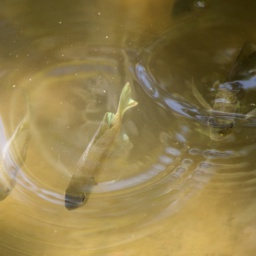}
		 & \includegraphics[height=\turnheightnew]{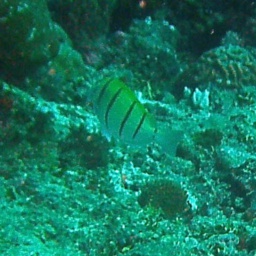}
		 & \includegraphics[height=\turnheightnew]{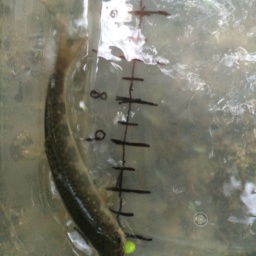}
		 & \includegraphics[height=\turnheightnew]{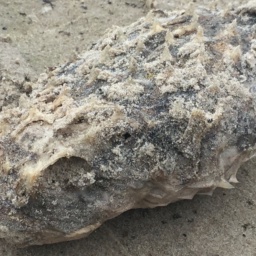}
		 & \includegraphics[height=\turnheightnew]{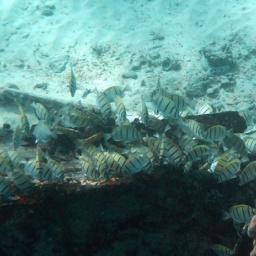}
		 & \includegraphics[height=\turnheightnew]{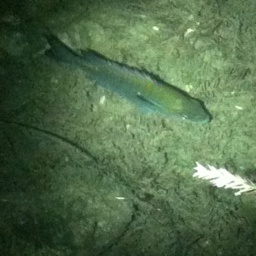}
		 & \includegraphics[height=\turnheightnew]{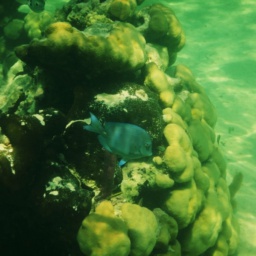}
		 & \includegraphics[height=\turnheightnew]{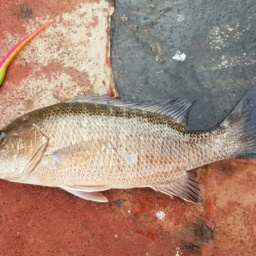}
		 & \includegraphics[height=\turnheightnew]{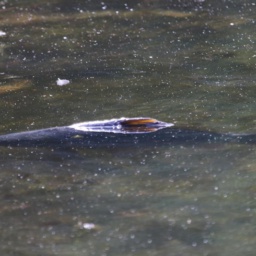}
		 & \includegraphics[height=\turnheightnew]{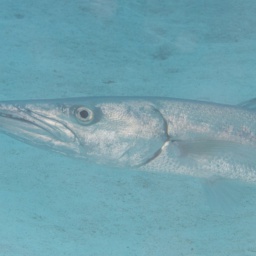}\\
		{\rotatebox{90}{\hspace{8pt}Amphib}} 
		 & \includegraphics[height=\turnheightnew]{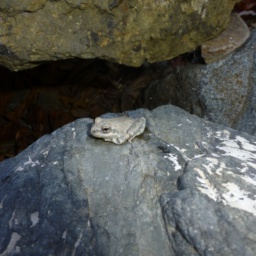}
		 & \includegraphics[height=\turnheightnew]{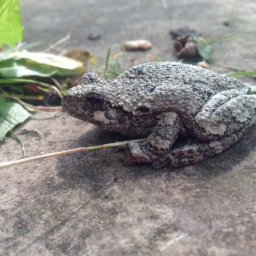}
		 & \includegraphics[height=\turnheightnew]{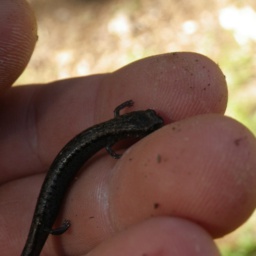}
		 & \includegraphics[height=\turnheightnew]{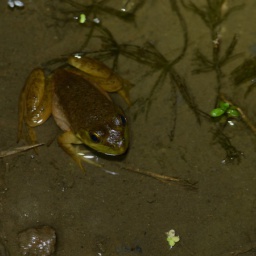}
		 & \includegraphics[height=\turnheightnew]{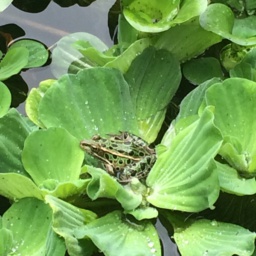}
		 & \includegraphics[height=\turnheightnew]{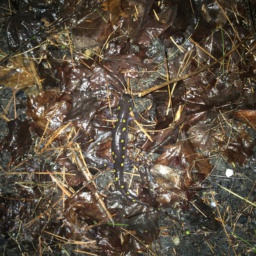}
		 & \includegraphics[height=\turnheightnew]{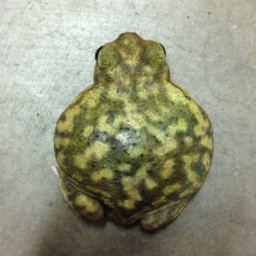}
		 & \includegraphics[height=\turnheightnew]{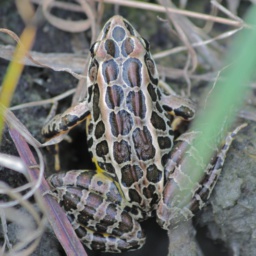}
		 & \includegraphics[height=\turnheightnew]{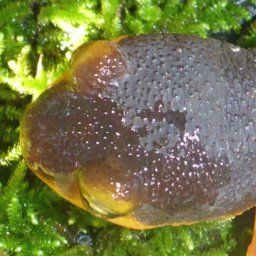}
		 & \includegraphics[height=\turnheightnew]{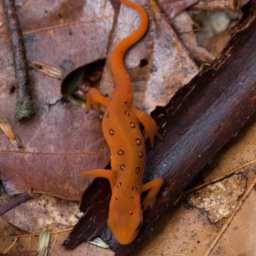}\\
		{\rotatebox{90}{\hspace{8pt}Animal}} 
		 & \includegraphics[height=\turnheightnew]{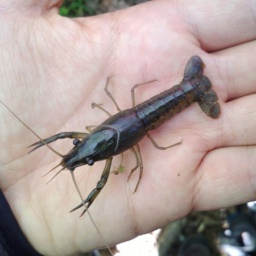}
		 & \includegraphics[height=\turnheightnew]{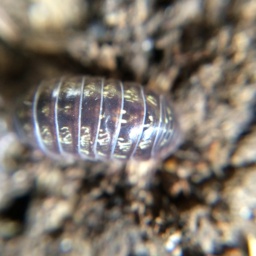}
		 & \includegraphics[height=\turnheightnew]{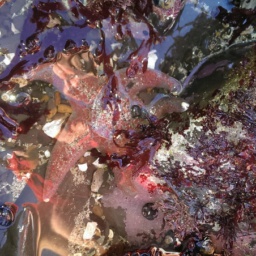}
		 & \includegraphics[height=\turnheightnew]{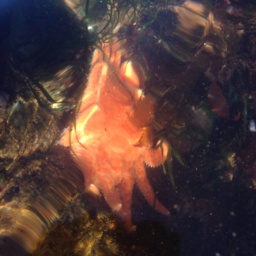}
		 & \includegraphics[height=\turnheightnew]{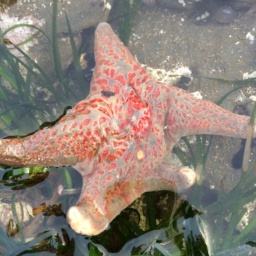}
		 & \includegraphics[height=\turnheightnew]{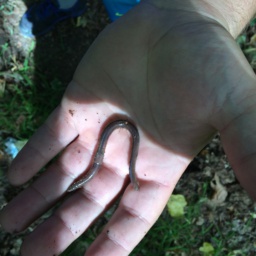}
		 & \includegraphics[height=\turnheightnew]{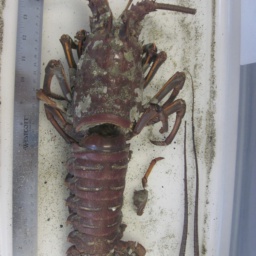}
		 & \includegraphics[height=\turnheightnew]{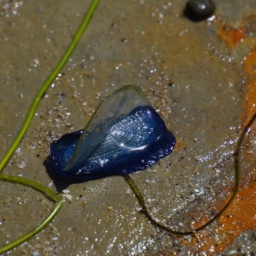}
		 & \includegraphics[height=\turnheightnew]{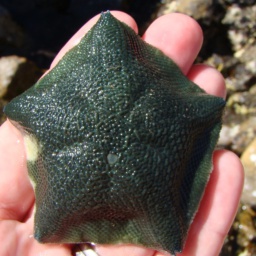}
		 & \includegraphics[height=\turnheightnew]{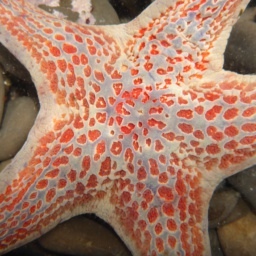}\\
		{\rotatebox{90}{\hspace{8pt}Arachn}} 
		 & \includegraphics[height=\turnheightnew]{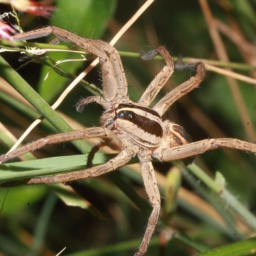}
		 & \includegraphics[height=\turnheightnew]{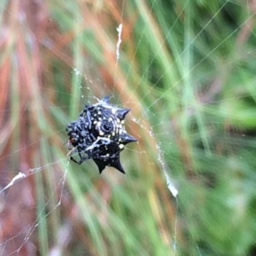}
		 & \includegraphics[height=\turnheightnew]{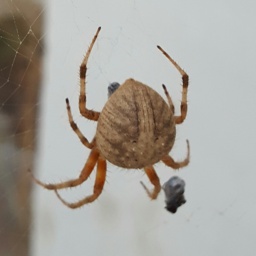}
		 & \includegraphics[height=\turnheightnew]{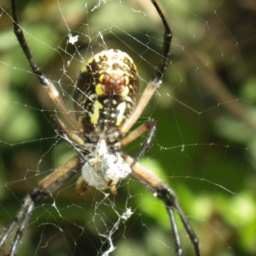}
		 & \includegraphics[height=\turnheightnew]{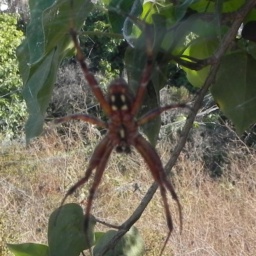}
		 & \includegraphics[height=\turnheightnew]{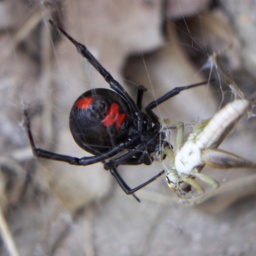}
		 & \includegraphics[height=\turnheightnew]{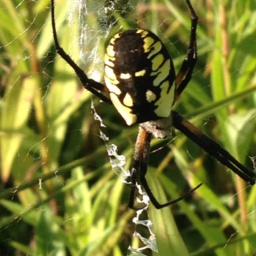}
		 & \includegraphics[height=\turnheightnew]{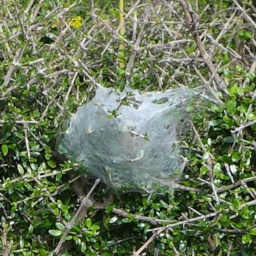}
		 & \includegraphics[height=\turnheightnew]{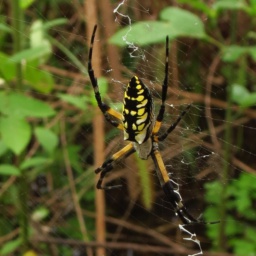}
		 & \includegraphics[height=\turnheightnew]{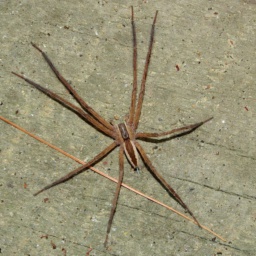}\\
		{\rotatebox{90}{\hspace{8pt}Aves}} 
		 & \includegraphics[height=\turnheightnew]{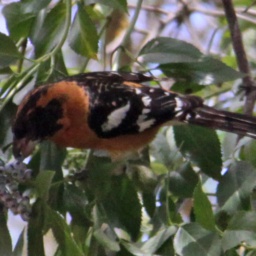}
		 & \includegraphics[height=\turnheightnew]{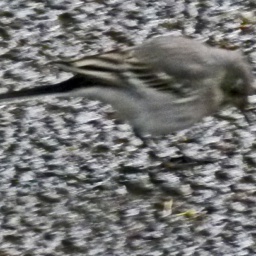}
		 & \includegraphics[height=\turnheightnew]{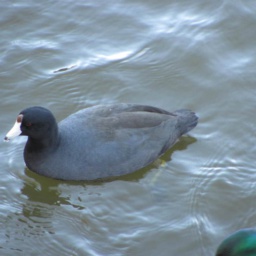}
		 & \includegraphics[height=\turnheightnew]{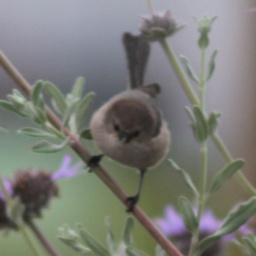}
		 & \includegraphics[height=\turnheightnew]{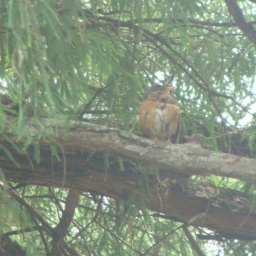}
		 & \includegraphics[height=\turnheightnew]{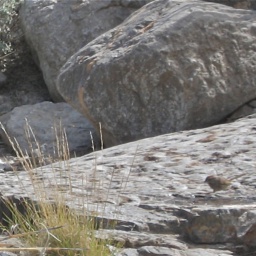}
		 & \includegraphics[height=\turnheightnew]{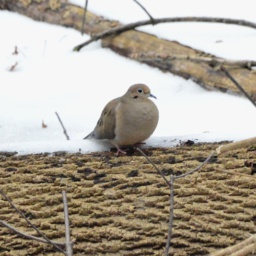}
		 & \includegraphics[height=\turnheightnew]{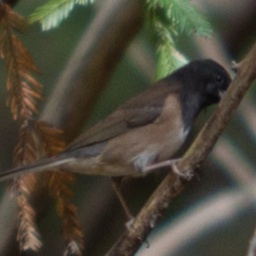}
		 & \includegraphics[height=\turnheightnew]{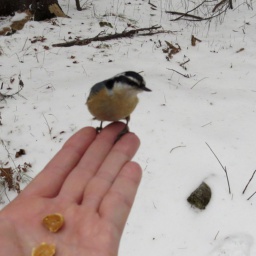}
		 & \includegraphics[height=\turnheightnew]{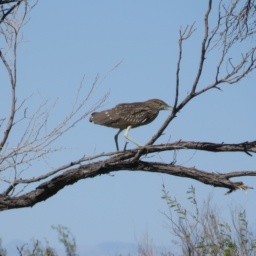}\\
		{\rotatebox{90}{\hspace{8pt}Chromi}} 
		 & \includegraphics[height=\turnheightnew]{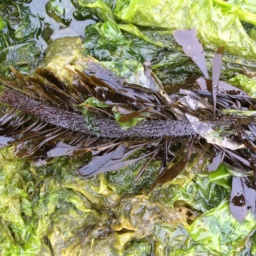}
		 & \includegraphics[height=\turnheightnew]{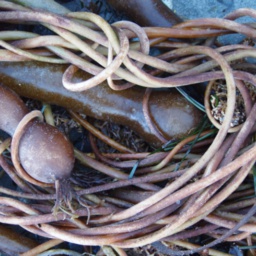}
		 & \includegraphics[height=\turnheightnew]{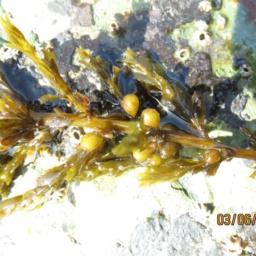}
		 & \includegraphics[height=\turnheightnew]{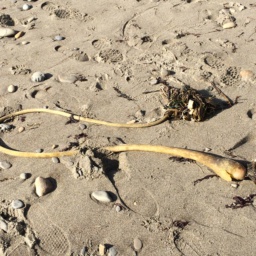}
		 & \includegraphics[height=\turnheightnew]{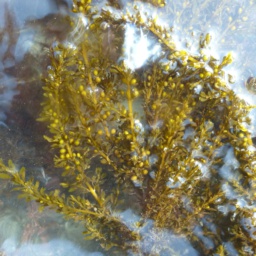}
		 & \includegraphics[height=\turnheightnew]{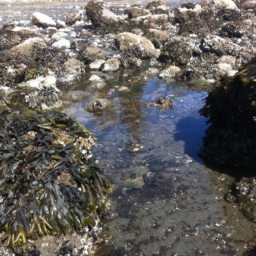}
		 & \includegraphics[height=\turnheightnew]{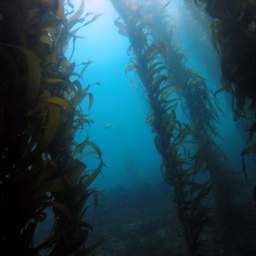}
		 & \includegraphics[height=\turnheightnew]{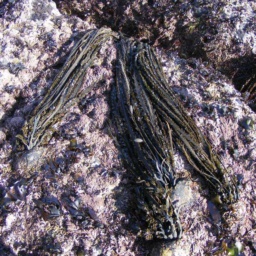}
		 & \includegraphics[height=\turnheightnew]{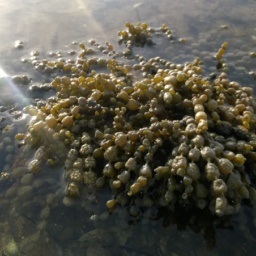}
		 & \includegraphics[height=\turnheightnew]{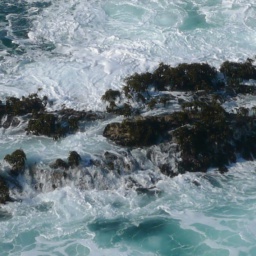}\\
		{\rotatebox{90}{\hspace{8pt}Fungi}} 
		 & \includegraphics[height=\turnheightnew]{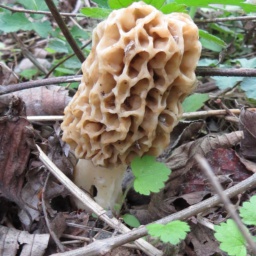}
		 & \includegraphics[height=\turnheightnew]{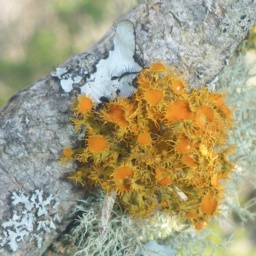}
		 & \includegraphics[height=\turnheightnew]{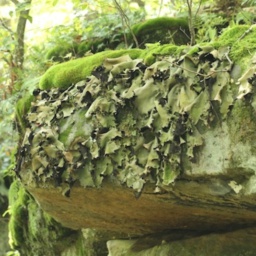}
		 & \includegraphics[height=\turnheightnew]{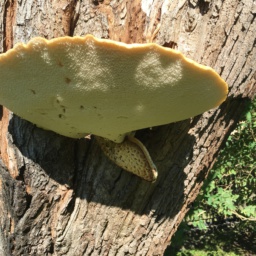}
		 & \includegraphics[height=\turnheightnew]{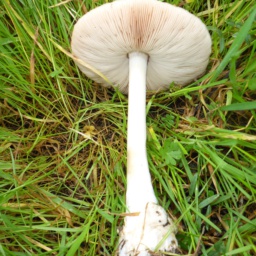}
		 & \includegraphics[height=\turnheightnew]{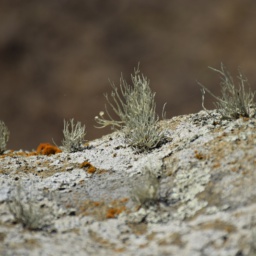}
		 & \includegraphics[height=\turnheightnew]{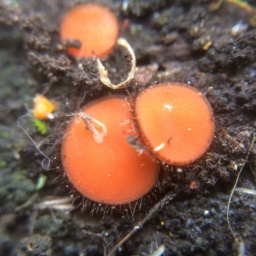}
		 & \includegraphics[height=\turnheightnew]{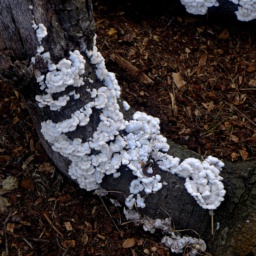}
		 & \includegraphics[height=\turnheightnew]{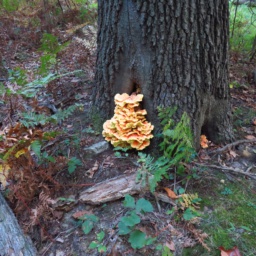}
		 & \includegraphics[height=\turnheightnew]{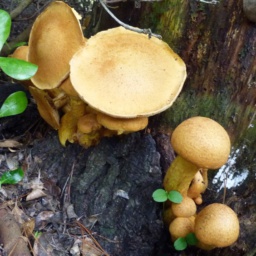}\\
		{\rotatebox{90}{\hspace{8pt}Insect}} 
		 & \includegraphics[height=\turnheightnew]{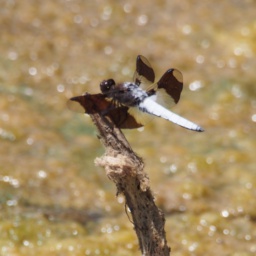}
		 & \includegraphics[height=\turnheightnew]{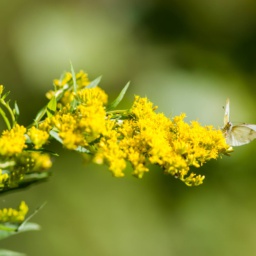}
		 & \includegraphics[height=\turnheightnew]{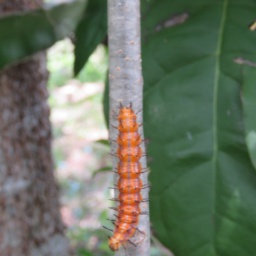}
		 & \includegraphics[height=\turnheightnew]{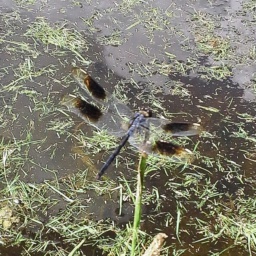}
		 & \includegraphics[height=\turnheightnew]{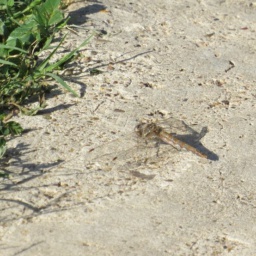}
		 & \includegraphics[height=\turnheightnew]{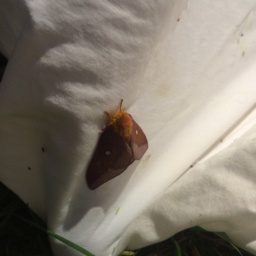}
		 & \includegraphics[height=\turnheightnew]{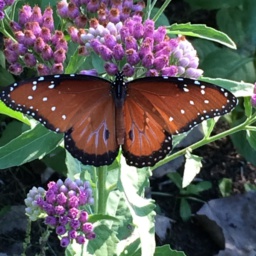}
		 & \includegraphics[height=\turnheightnew]{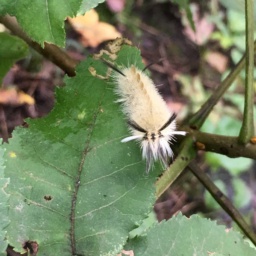}
		 & \includegraphics[height=\turnheightnew]{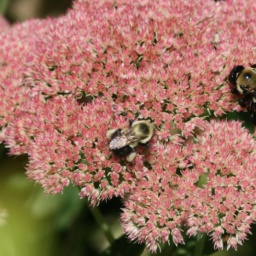}
		 & \includegraphics[height=\turnheightnew]{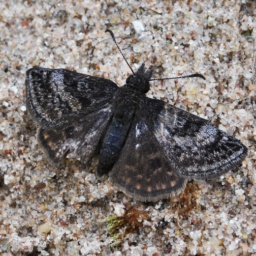}\\
		{\rotatebox{90}{\hspace{8pt}Mammal}} 
		 & \includegraphics[height=\turnheightnew]{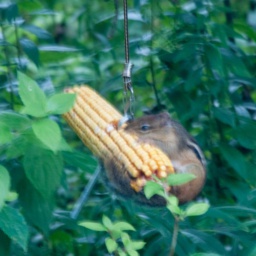}
		 & \includegraphics[height=\turnheightnew]{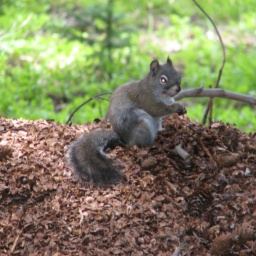}
		 & \includegraphics[height=\turnheightnew]{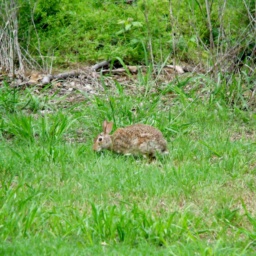}
		 & \includegraphics[height=\turnheightnew]{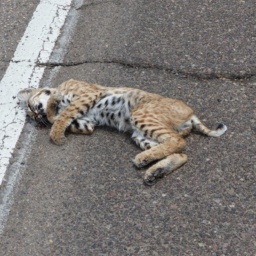}
		 & \includegraphics[height=\turnheightnew]{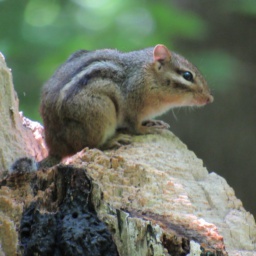}
		 & \includegraphics[height=\turnheightnew]{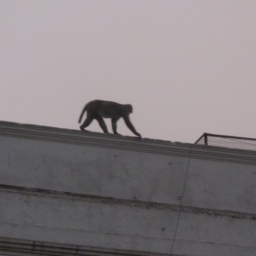}
		 & \includegraphics[height=\turnheightnew]{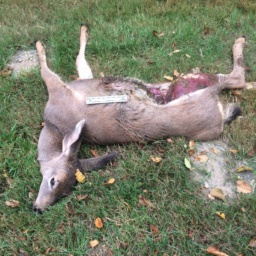}
		 & \includegraphics[height=\turnheightnew]{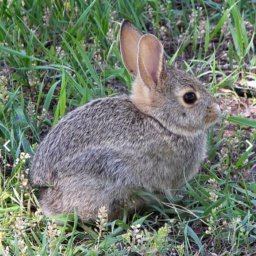}
		 & \includegraphics[height=\turnheightnew]{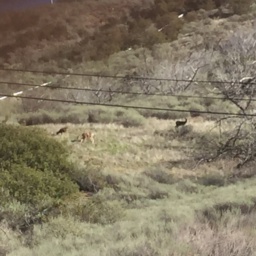}
		 & \includegraphics[height=\turnheightnew]{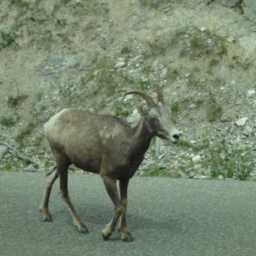}\\
		{\rotatebox{90}{\hspace{8pt}Mollus}} 
		 & \includegraphics[height=\turnheightnew]{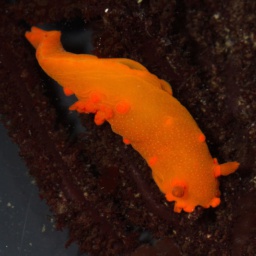}
		 & \includegraphics[height=\turnheightnew]{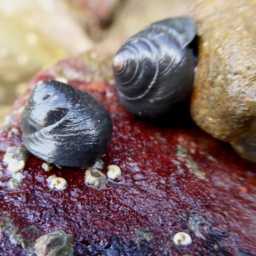}
		 & \includegraphics[height=\turnheightnew]{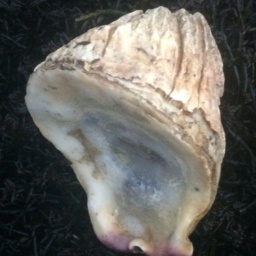}
		 & \includegraphics[height=\turnheightnew]{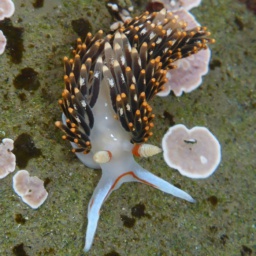}
		 & \includegraphics[height=\turnheightnew]{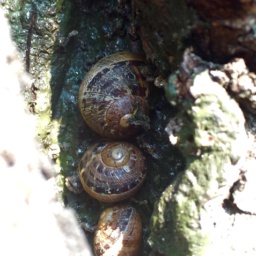}
		 & \includegraphics[height=\turnheightnew]{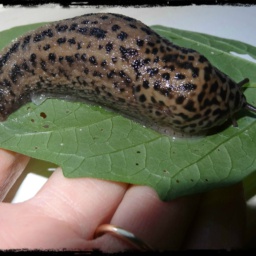}
		 & \includegraphics[height=\turnheightnew]{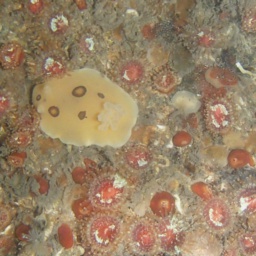}
		 & \includegraphics[height=\turnheightnew]{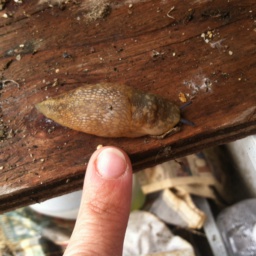}
		 & \includegraphics[height=\turnheightnew]{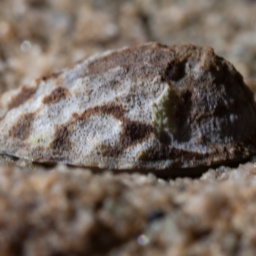}
		 & \includegraphics[height=\turnheightnew]{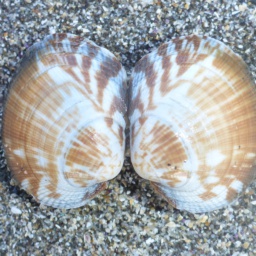}\\
		{\rotatebox{90}{\hspace{8pt}Planta}} 
		 & \includegraphics[height=\turnheightnew]{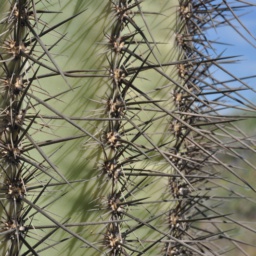}
		 & \includegraphics[height=\turnheightnew]{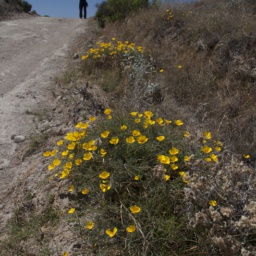}
		 & \includegraphics[height=\turnheightnew]{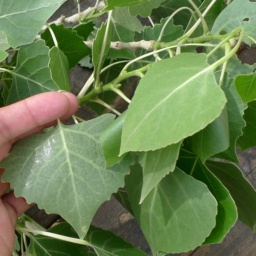}
		 & \includegraphics[height=\turnheightnew]{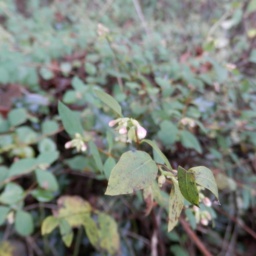}
		 & \includegraphics[height=\turnheightnew]{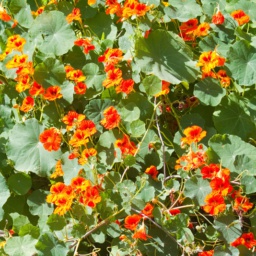}
		 & \includegraphics[height=\turnheightnew]{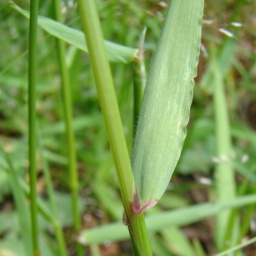}
		 & \includegraphics[height=\turnheightnew]{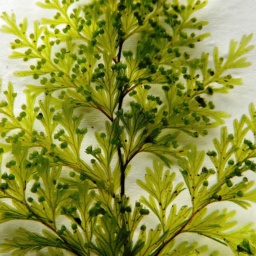}
		 & \includegraphics[height=\turnheightnew]{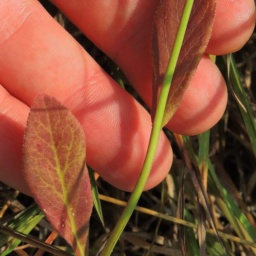}
		 & \includegraphics[height=\turnheightnew]{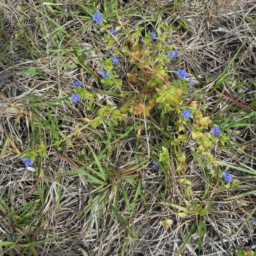}
		 & \includegraphics[height=\turnheightnew]{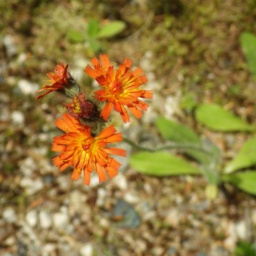}\\
		{\rotatebox{90}{\hspace{8pt}Protoz}} 
		 & \includegraphics[height=\turnheightnew]{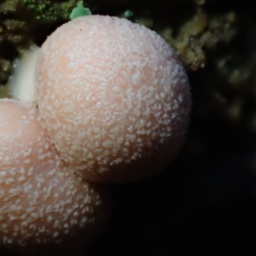}
		 & \includegraphics[height=\turnheightnew]{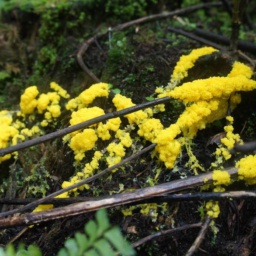}
		 & \includegraphics[height=\turnheightnew]{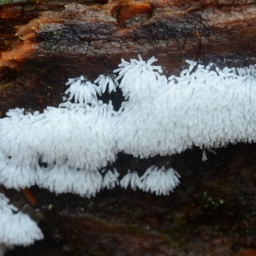}
		 & \includegraphics[height=\turnheightnew]{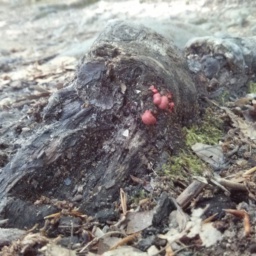}
		 & \includegraphics[height=\turnheightnew]{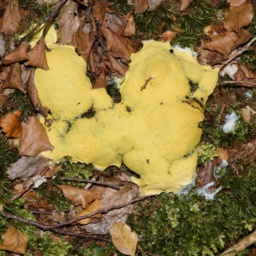}
		 & \includegraphics[height=\turnheightnew]{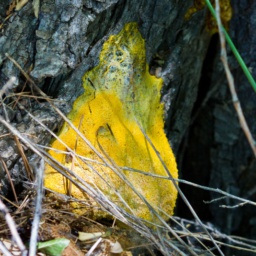}
		 & \includegraphics[height=\turnheightnew]{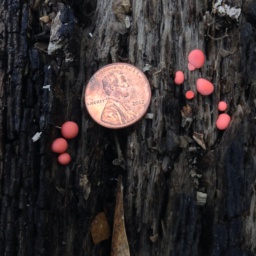}
		 & \includegraphics[height=\turnheightnew]{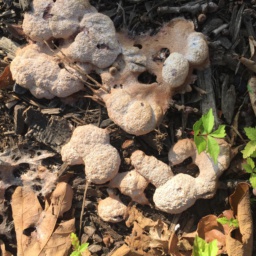}
		 & \includegraphics[height=\turnheightnew]{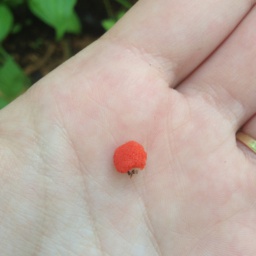}
		 & \includegraphics[height=\turnheightnew]{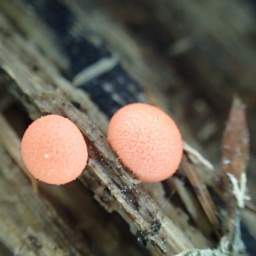}\\
		{\rotatebox{90}{\hspace{8pt}Reptil}} 
		 & \includegraphics[height=\turnheightnew]{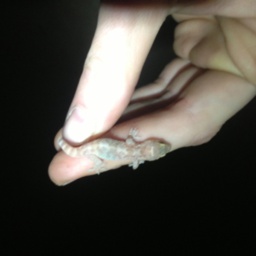}
		 & \includegraphics[height=\turnheightnew]{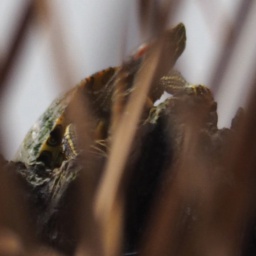}
		 & \includegraphics[height=\turnheightnew]{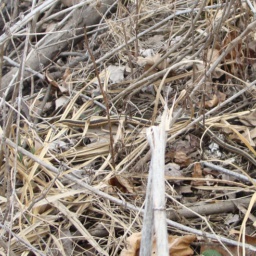}
		 & \includegraphics[height=\turnheightnew]{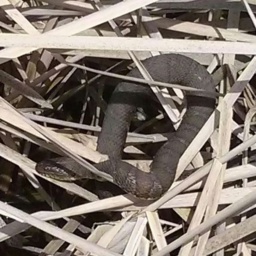}
		 & \includegraphics[height=\turnheightnew]{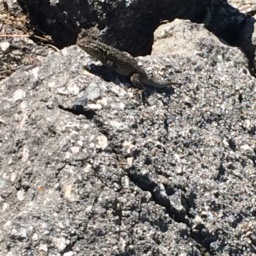}
		 & \includegraphics[height=\turnheightnew]{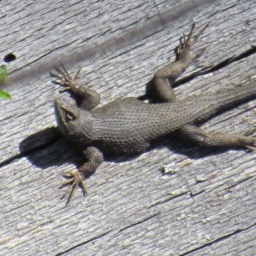}
		 & \includegraphics[height=\turnheightnew]{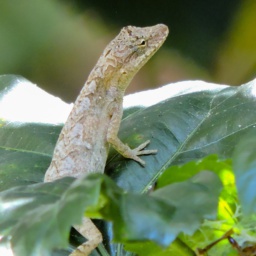}
		 & \includegraphics[height=\turnheightnew]{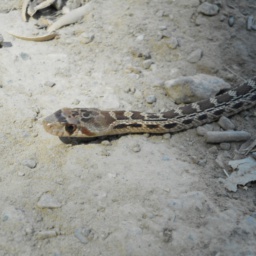}
		 & \includegraphics[height=\turnheightnew]{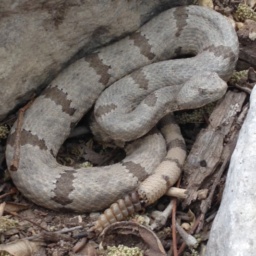}
		 & \includegraphics[height=\turnheightnew]{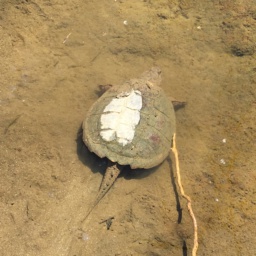}\\
\end{tabular}